\documentclass{article}
\PassOptionsToPackage{numbers, compress}{natbib}
\usepackage[final]{neurips_2024}

\usepackage{verbatim}
\usepackage[pdftex]{graphicx}
\usepackage[hyphens]{url}
\usepackage[hidelinks,colorlinks=true,linkcolor=blue,citecolor=blue]{hyperref}       
\usepackage{booktabs}       
\usepackage{amsmath, amsfonts}       
\usepackage{nicefrac}       
\usepackage{microtype}      
\usepackage{caption}
\usepackage{subcaption}
\usepackage[hyphens]{url}
\usepackage{cleveref}
\usepackage{algorithm}
\usepackage{algpseudocode}

\hypersetup{
    linkcolor=blue,
    citecolor=blue,      
    urlcolor=blue,
    pdfpagemode=FullScreen,
}

\usepackage{url}

\usepackage{amsmath,amsthm,amssymb,amsbsy,amsfonts,amscd,bm}
\usepackage{paralist}
\usepackage{xcolor}
\usepackage{color}
\usepackage{graphicx}
\graphicspath{{./figs/}}
\usepackage{comment}
\usepackage{multirow}
\usepackage{enumitem}
\usepackage{fancyhdr}
\usepackage{subcaption}
\usepackage{wrapfig}

\newtheorem{thm}{Theorem}
\newtheorem{assum}{Assumption}

\newcommand{\e}{\begin{equation}}
\newcommand{\ee}{\end{equation}}
\newcommand{\en}{\begin{equation*}}
\newcommand{\een}{\end{equation*}}
\newcommand{\eqn}{\begin{eqnarray}}
\newcommand{\eeqn}{\end{eqnarray}}
\newcommand{\bmat}{\begin{bmatrix}}
\newcommand{\emat}{\end{bmatrix}}

\DeclareMathAlphabet\mathbfcal{OMS}{cmsy}{b}{n}

\newcommand{\mb}{\bm}
\newcommand{\mc}{\mathcal}

\newcommand{\bb}{\mathbb}

\newcommand{\diag}{\operatorname{diag}}

\DeclareMathOperator*{\argmin}{\text{arg~min}}

\newcommand{\norm}[2]{\left\| #1 \right\|_{#2}}

\newcommand{\abs}[1]{\left| #1 \right|}

\newcommand{\innerprod}[2]{\left\langle #1,  #2 \right\rangle}

\newcommand{ \brac }[1]{\left[ #1 \right]}
\newcommand{ \paren }[1]{ \left( #1 \right) }

\graphicspath{{./figs/}}

\newlength{\imgwidth}
\newboolean{twoColVersion}
\setboolean{twoColVersion}{false}
\newcommand{\twoCol}[2]{\ifthenelse{\boolean{twoColVersion}} {#1} {#2} }

\title{Understanding Generalizability of Diffusion Models Requires Rethinking the Hidden Gaussian Structure}
\author{%
  Xiang Li\textsuperscript{1}, Yixiang Dai\textsuperscript{1}, Qing Qu\textsuperscript{1}\\
  \textsuperscript{1}Department of EECS, University of Michigan,\\
  \href{mailto:forkobe@umich.edu}{forkobe@umich.edu}, \href{mailto:yixiang@umich.edu}{yixiang@umich.edu},  
  \href{mailto:qingqu@umich.edu}{qingqu@umich.edu}
}
\pdfoutput=1

\begin{document}

\maketitle
\vspace{-0.1in}
\begin{abstract}
\vspace{-0.1in}
In this work, we study the generalizability of diffusion models by looking into the hidden properties of the learned score functions, which are essentially a series of deep denoisers trained on various noise levels. We observe that as diffusion models transition from memorization to generalization, their corresponding nonlinear diffusion denoisers exhibit increasing linearity. This discovery leads us to investigate the linear counterparts of the nonlinear diffusion models, which are a series of linear models trained to match the function mappings of the nonlinear diffusion denoisers. Interestingly, these linear denoisers are nearly optimal for multivariate Gaussian distributions defined by the empirical mean and covariance of the training dataset, and they effectively approximate the behavior of nonlinear diffusion models. This finding implies that diffusion models have the inductive bias towards capturing and utilizing the Gaussian structure (covariance information) of the training dataset for data generation. We empirically demonstrate that this inductive bias is a unique property of diffusion models in the generalization regime, which becomes increasingly evident when the model's capacity is relatively small compared to the training dataset size. In the case where the model is highly overparameterized, this inductive bias emerges during the initial training phases before the model fully memorizes its training data. Our study provides crucial insights into understanding the notable strong generalization phenomenon recently observed in real-world diffusion models.

\end{abstract}

\vspace{-0.1in}
\section{Introduction}
\vspace{-0.1in}

In recent years, diffusion models~\cite{sohl2015deep, ho2020denoising, songscore, karras2022elucidating} have become one of the leading generative models, powering the state-of-the-art image generation systems such as Stable Diffusion~\cite{rombach2022high}. To understand the empirical success of diffusion models, several works~\cite{chen2023sampling, de2022convergence, lee2022convergence,lee2023convergence,wu2024stochastic,li2024sharp,huang2024denoising} have focused on their sampling behavior, showing that the data distribution can be effectively estimated in the reverse sampling process, assuming that the score function is learned accurately. Meanwhile, other works~\cite{chen2023score,oko2023diffusion,shah2023learning,cui2023analysis,zhang2023emergence,wang2024diffusion} investigate the learning of score functions, showing that effective approximation can be achieved with score matching loss under certain assumptions. However, these theoretical insights, grounded in simplified assumptions about data distribution and neural network architectures, do not fully capture the complex dynamics of diffusion models in practical scenarios. One significant discrepancy between theory and practice is that real-world diffusion models are trained only on a finite number of data points. As argued in~\cite{li2024good}, theoretically a perfectly learned score function over the empirical data distribution can only replicate the training data.
In contrast, diffusion models trained on finite samples exhibit remarkable generalizability, producing high-quality images that significantly differ from the training examples. Therefore, a good understanding of the remarkable generative power of diffusion models is still lacking. 

In this work, we aim to deepen the understanding of generalizability in diffusion models by analyzing the inherent properties of the learned score functions. 
Essentially, the score functions can be interpreted as a series of deep denoisers trained on various noise levels. These denoisers are then chained together to progressively denoise a randomly sampled Gaussian noise into its corresponding clean image, thus, understanding the function mappings of these diffusion denoisers is critical to demystify the working mechanism of diffusion models. Motivated by the linearity observed in the diffusion denoisers of effectively generalized diffusion models, we propose to elucidate their function mappings with a linear distillation approach, where the resulting linear models serve as the linear approximations of their nonlinear counterparts.
\vspace{-0.1in}
\paragraph{Contributions of this work:} Our key findings can be highlighted as follows:
\begin{itemize}[leftmargin=*]
    \item \textbf{Inductive bias towards Gaussian structures (\Cref{sec: Inductive bias towards learning Gaussian}).} Diffusion models in the \emph{generalization regime} exhibit an inductive bias towards learning diffusion denoisers that are close (but not equal) to the optimal denoisers for a multivariate Gaussian distribution, defined by the empirical mean and covariance of the training data.
    This implies the diffusion models have the inductive bias towards capturing the Gaussian structure (covariance information) of the training data for image generation. 

    \item \textbf{Model Capacity and Training Duration (\Cref{sec: conditon for the inductive bias})} We show that this inductive bias is most pronounced when the model capacity is relatively small compared to the size of the training data. However, even if the model is highly overparameterized, such inductive bias still emerges during early training phases, before the model memorizes its training data. This implies that early stopping can prompt generalization in overparameterized diffusion models.
    
    \item \textbf{Connection between Strong Generalization and Gaussian Structure (\Cref{strong generalizability}).} Lastly, we argue that the recently observed strong generalization~\cite{kadkhodaie2023generalization} results from diffusion models learning certain common low-dimensional structural features shared across non-overlapping datasets. We show that such low-dimensional features can be partially explained through the Gaussian structure.
\end{itemize}

\paragraph{Relationship with Prior Arts.} Recent research~\cite{kadkhodaie2023generalization,somepalli2023diffusion,somepalli2023understanding,yoon2023diffusion,gu2023memorization} demonstrates that diffusion models operate in two distinct regimes: (\emph{i}) a memorization regime, where models primarily reproduce training samples and (\emph{ii}) a generalization regime, where models generate high-quality, novel images that extend beyond the training data.
In the generalization regime, a particularly intriguing phenomenon is that diffusion models trained on non-overlapping datasets can generate nearly identical samples ~\cite{kadkhodaie2023generalization}. While prior work~\cite{kadkhodaie2023generalization} attributes this "\emph{strong generalization}" effect to the structural inductive bias inherent in diffusion models leading to the optimal denoising basis (geometry-adaptive harmonic basis), our research advances this understanding by demonstrating diffusion models' inductive bias towards capturing the Gaussian structure of the training data. Our findings also corroborate with observations of earlier study~\cite{wang2023hidden} that the learned score functions of well-trained diffusion models closely align with the optimal score functions of a multivariate Gaussian approximation of the training data.

\vspace{-0.1in}
\section{Preliminary}
\label{preliminary}
\vspace{-0.1in}

\paragraph{Basics of Diffusion Models.} 
Given a data distribution $p_{\text{data}}(\mb x)$, where $\mb x \in \bb R^d$, diffusion models~\cite{sohl2015deep, ho2020denoising, songscore, karras2022elucidating} define a series of intermediate states $p(\mb x;\sigma(t))$ by adding Gaussian noise sampled from $\mathcal{N}(\mb 0,\sigma(t)^2\mb I)$ to the data, where $\sigma(t)$ is a predefined schedule that specifies the noise level at time $t \in [0,T]$, such that at the end stage the noise mollified distribution $p(\mb x;\sigma(T))$ is indistinguishable from the pure Gaussian distribution. Subsequently, a new sample is generated by progressively denoising a random noise $\mb x_T \sim \mathcal{N}(\mb 0, \sigma(T)^2\mb I)$ to its corresponding clean image $\mb x_0$. 

Following~\cite{karras2022elucidating}, this forward and backward diffusion process can be expressed with a probabilistic ODE:
\begin{align}\label{eqn:ode}
    d\mb x = -\dot \sigma(t)\sigma(t)\nabla_{\mb x}\log p(\mb x;\sigma(t))dt.
\end{align}
In practice the score function $\nabla_{\mb x}\log p(\mb x;\sigma(t))$ can be approximated by
\begin{align}
    \nabla_{\mb x}\log p(\mb x;\sigma(t))=(\mathcal D_{\mb \theta}(\mb x;\sigma(t))-\mb x)/\sigma(t)^2,
\end{align}
where $\mathcal D_{\mb \theta}(\mb x;\sigma (t))$ is parameterized by a deep network with parameters $\mb \theta$ trained with the denoising score matching objective: 
\begin{align} \label{Training Denoisers}
   \min_{\mb \theta} \mathbb{E}_{\mb x \sim p_{\text{data}}}\mathbb{E}_{\mb \epsilon \sim \mathcal{N}(\mb 0, \sigma(t)^2\mb I)} \brac{ \|\mathcal D_{\mb \theta}(\mb x+\mb \epsilon;\sigma (t))-\mb x\|_2^2}.
\end{align}

In the discrete setting, the reverse ODE in \eqref{eqn:ode} takes the following form:
\begin{align}
    \label{1st order update 3}
    \mb x_{i+1}\leftarrow (1-(t_{i}-t_{i+1})\frac{\dot\sigma(t_i)}{\sigma(t_i)})\mb x_i+(t_i-t_{i+1})\frac{\dot\sigma(t_i)}{\sigma(t_i)}\mathcal D_{\mb \theta}(\mb x_i;\sigma(t_i)),
\end{align}
where $\mb x_0\sim \mathcal{N}(\mb 0,\sigma^2(t_0)\mb I)$.
Notice that at each iteration $i$, the intermediate sample $\mb x_{i+1}$ is the sum of the scaled $\mb x_i$ and the denoising output $\mathcal D_{\mb \theta}(\mb x_i;\sigma(t_i))$. Obviously, the final sampled image is largely determined by the denoiser $ \mathcal D_{\mb \theta}(\mb x;\sigma(t))$. If we can understand the function mapping of these diffusion denoisers, we can demystify the working mechanism of diffusion models.


\vspace{-0.1in}
\paragraph{Optimal Diffusion Denoisers under Simplified Data Assumptions.}
Under certain assumptions on the data distribution $p_{\text{data}}(\mb x)$, the optimal diffusion denoisers $ \mathcal D_{\mb \theta}(\mb x;\sigma(t))$ that minimize the score matching objective~\eqref{Training Denoisers} can be derived analytically in closed-forms as we discuss below.
\begin{itemize}[leftmargin=*]
    \item \textit{Multi-delta distribution of the training data.} Suppose the training dataset contains a finite number of data points $\{\mb y_1, \mb y_2,...,\mb y_N\}$, a natural way to model the data distribution is to represent it as a multi-delta distribution: $p(\mb x)=\frac{1}{N}\sum_{i=1}^N\delta(\mb x-\mb y_i)$. In this case, the optimal denoiser is
    \begin{align}
        \label{multi-delta denoiser}
        \mathcal D_{\mathrm{M}}(\mb x;\sigma(t))=\frac{\sum_{i=1}^{N}\mathcal{N}(\mb x;\mb y_i,\sigma(t)^2\mb I)\mb y_i}{\sum_{i=1}^{N}\mathcal{N}(\mb x;\mb y_i,\sigma(t)^2\mb I)},
    \end{align}
    which is essentially a softmax-weighted combination of the finite data points. As proved in~\cite{gu2023memorization}, such diffusion denoisers $\mathcal D_{\mathrm{M}}(\mb x;\sigma(t))$ can only generate exact replicas of the training samples, therefore they have no generalizability.
    
    \item \textit{Multivariate Gaussian distribution.} Recent work~\cite{wang2023hidden} suggests modeling the data distribution $p_{\text{data}}(\mb x)$ as a multivariate Gaussian distribution $p(\mb x)=\mathcal{N}(\mb \mu,\mb \Sigma)$, where the mean $\mb \mu$ and the covariance $\mb \Sigma$ are approximated by the empirical mean $\mb\mu=\frac{1}{N}\sum_{i=1}^N\mb y_i$ and the empirical covariance $\mb \Sigma=\frac{1}{N}\sum_{i=1}^{N}(\mb y_i-\mb \mu)(\mb y_i-\mb \mu)^T$ of the training dataset. In this case, the optimal denoiser is: 
    \begin{align}
      \label{Gaussian denoiser}
      \mathcal D_{\mathrm{G}}(\mb x;\sigma(t))=\mb \mu+\mb U\Tilde{\mb \Lambda}_{\sigma(t)}\mb U^T (\mb x-\mb \mu),
    \end{align}
     where $\mb \Sigma=\mb U \mb \Lambda\mb U^T$ is the SVD of the empirical covariance matrix, with singular values $\mb \Lambda = \diag \paren{ \lambda_1,\cdots,\lambda_d }$ and $\Tilde{\mb \Lambda}_{\sigma(t)}=\diag \paren{\frac{\lambda_1}{\lambda_1+\sigma(t)^2}, \cdots,
     \frac{\lambda_d}{\lambda_d+\sigma(t)^2} }$. With this linear Gaussian denoiser, as proved in~\cite{wang2023hidden}, the sampling trajectory of the probabilistic ODE~\eqref{eqn:ode} has close form:
     \begin{align}
        \label{Gaussian ODE}
         \mb x_t = \mb \mu + \sum_{i=1}^{d}\sqrt{\frac{\sigma(t)^2+\lambda_i}{\sigma(T)^2+\lambda_i}}\mb u_i^T(\mb x_T-\mb \mu)\mb u_i,
     \end{align}
     where $\mb u_i$ is the $i^{\text{th}}$ singular vector of the empirical covariance matrix. 
    While~\cite{wang2023hidden} demonstrate that the Gaussian scores approximate learned scores at high noise variances, we show that they are nearly the best linear approximations of learned scores across a much wider range of noise variances.
\end{itemize}
\vspace{-0.1in}
\paragraph{Generalization vs. Memorization of Diffusion Models.} As the training dataset size increases, diffusion models transition from the memorization regime—where they can only replicate its training images---to the generalization regime, where the they produce high-quality, novel images \cite{zhang2023emergence}. While memorization can be interpreted as an overfitting of diffusion models to the training samples, the mechanisms underlying the generalization regime remain less well understood. This study aims to explore and elucidate the inductive bias that enables effective generalization in diffusion models.

\vspace{-0.1in}
\section{Hidden Linear and Gaussian Structures in Diffusion Models}
\vspace{-0.1in}
\label{sec: Inductive bias towards learning Gaussian}

In this section, we study the intrinsic structures of the learned score functions of diffusion models in the generalization regime.
Through various experiments and theoretical investigation, we show that
\begin{center}
\emph{\textbf{Diffusion models in the generalization regime have inductive bias towards learning the Gaussian structures of the dataset.}}
\end{center}

Based on the \emph{linearity} observed in diffusion denoisers trained in the generalization regime, we propose to investigate their intrinsic properties through a \emph{linear distillation} technique, with which we train a series of linear models to approximate the nonlinear diffusion denoisers (\Cref{linear characteristic}). Interestingly, these linear models closely resemble the optimal denoisers for a multivariate Gaussian distribution characterized by the empirical mean and covariance of the training dataset (\Cref{inductive bias towards Gaussian}). This implies diffusion models have the inductive bias towards learning the Gaussian structure of the training dataset. We theoretically show that the observed Gaussian structure is the optimal solution to the denoising score matching objective under the constraint that the model is linear (\Cref{theoretical analysis}). In the subsequent sections, although we mainly demonstrate our results using the FFHQ datasets, our findings are robust and extend to various architectures and datasets, as detailed in~\Cref{Additional experiments}.

\vspace{-0.1in}
\subsection{Diffusion Models Exhibit Linearity in the Generalization Regime}
\vspace{-0.1in}
\label{linear characteristic}
 Our study is motivated by the emerging linearity observed in diffusion models in the generalization regime. Specifically, we quantify the linearity of diffusion denoisers at various noise level $\sigma(t)$ by jointly assessing their "Additivity" and "Homogeneity" with a linearity score (LS) defined by the cosine similarity between $\mathcal D_{\mb \theta} (\alpha \mb x_1+\beta \mb x_2; \sigma (t))$ and $\alpha \mathcal D_{\mb \theta}(\mb x_1;\sigma(t))+\beta \mathcal D_{\mb\theta}(\mb x_2;\sigma(t))$:
\begin{align*}
  \texttt{LS}(t) \;=\; \bb E_{ \mb x_1, \mb x_2 \sim p(\mb x;\sigma(t)) }\brac{ \abs{\innerprod{ \frac{\mathcal D_{\mb \theta} (\alpha \mb x_1+\beta \mb x_2; \sigma (t))}{ \norm{\mathcal D_{\mb \theta} (\alpha \mb x_1+\beta \mb x_2; \sigma (t))}{2} }  }{ \frac{ \alpha \mathcal D_{\mb \theta}(\mb x_1;\sigma(t))+\beta \mathcal D_{\mb\theta}(\mb x_2;\sigma(t)) }{ \norm{\alpha \mathcal D_{\mb \theta}(\mb x_1;\sigma(t))+\beta \mathcal D_{\mb\theta}(\mb x_2;\sigma(t)) }{2} }} } },
\end{align*}
where $\mb x_1,\mb x_2 \sim p(\mb x;\sigma(t))$, and $\alpha \in \bb R$ and $\beta\in \bb R$ are scalars. In practice, the expectation is approximated with its empirical mean over 100 samples. A more detailed discussion on this choice of measuring linearity is deferred to~\Cref{sec: linearity of diffusion denoisers}. 

\begin{wrapfigure}[17]{r}{0.5\textwidth}
    \vspace{-1cm}
  \begin{center}
    \includegraphics[width=0.5\textwidth]{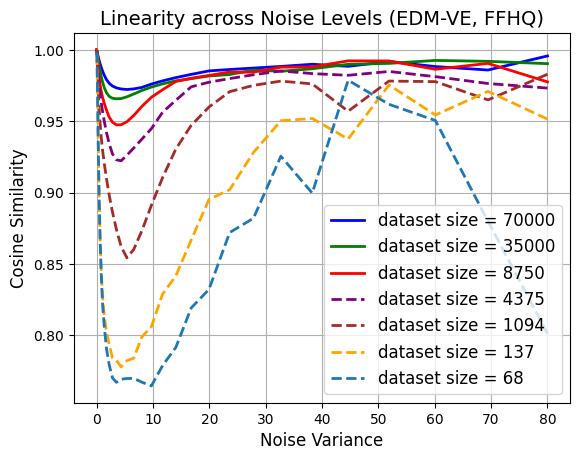}
  \end{center}
  \vspace{-0.3cm}
  \caption{\textbf{Linearity scores of diffusion denoisers.} Solid and dashed lines depict the linearity scores across noise variances for models in the generalization and memorization regimes, respectively, where $\alpha=\beta=1/\sqrt{2}$. }
  \label{fig:linearity}
\end{wrapfigure}
Following the EDM training configuration~\cite{karras2022elucidating}, we set the noise levels $\sigma(t)$ within the continuous range [0.002,80]. As shown in~\Cref{fig:linearity}, as diffusion models transition from the memorization regime to the generalization regime (increasing the training dataset size), the corresponding diffusion denoisers $\mathcal D_{\mb\theta}$ exhibit increasing linearity. This phenomenon persists across diverse datasets\footnote{For example, FFHQ~\cite{karras2019style}, CIFAR-10~\cite{krizhevsky2009learning}, AFHQ~\cite{choi2020stargan} and LSUN-Churches~\cite{yu2015lsun}.} as well as various training configurations\footnote{For example, EDM-VE, EDM-VP and EDM-ADM.}; see~\Cref{emerging linearity from memorization to generalization} for more details. This emerging linearity motivates us to ask the following questions:
\begin{center}
\begin{itemize}[leftmargin=*]
    \item \textit{To what extent can a diffusion model be approximated by a linear model?}
    \item \textit{If diffusion models can be approximated linearly, what are the underlying characteristics of this linear approximation?}
\end{itemize}
\end{center}

\vspace{-0.1in}
\paragraph{Investigating the Linear Structures via Linear Distillation.} To address these questions, we investigate the hidden linear structure of diffusion denoisers through \textit{linear distillation}. Specifically, for a given diffusion denoiser $\mathcal D_{\mb \theta}(\mb x;\sigma(t))$ at noise level $\sigma(t)$, we approximate it with a linear function (with a bias term) such that:
\begin{align}
    \label{linear model with bias}
    \mathcal D_{\mathrm{L}}(\mb x;\sigma(t)) := \mb W_{\sigma(t)} \mb x+\mb b_{\sigma(t)}\;\approx\; \mathcal D_{\mb \theta} (\mb x;\sigma(t)),\; \forall \mb x\sim p(\mb x;\sigma(t)),
\end{align} 
where the weight $\mb W_{\sigma(t)}\in \mathbb{R}^{d\times d}$ and bias $\mb b_{\sigma(t)}\in\mathbb{R}^d$ are learned by solving the following optimization problem with gradient descent:\footnote{For the following, the input is the vectorized version of the noisy image and the expectation is approximated using finite samples of input-output pairs $(\mb x_i+\mb \epsilon_i, \mathcal D_{\mb \theta}(\mb x_i+\mb \epsilon,\sigma(t)))$ with $i=1,...,N$ (see distillation details in~\Cref{sec: linear distillation}).}
\begin{align}
    \label{eq:linear distillation objective}
    \min_{\mb W_{\sigma(t)}, \mb b_{\sigma(t)}}\mathbb{E}_{\mb x \sim p_{\text{data}}(\mb x)}\mathbb{E}_{\mb \epsilon\sim\mathcal{N}(\mb 0,\sigma(t)^2\mb I)}||\mb W_{\sigma(t)}(\mb x+\mb \epsilon)+\mb b_{\sigma(t)}-\mathcal D_{\mb \theta}(\mb x+\mb\epsilon;\sigma(t))||_2^2.
\end{align}
If these linear models effectively approximate the nonlinear diffusion denoisers, analyzing their weights can elucidate the generation mechanism. 

While diffusion models are trained on continuous noise variance levels within [0.002,80], we examine the 10 discrete sampling steps specified by the EDM schedule~\cite{karras2022elucidating}: [80.0,
 42.415,
 21.108,
 9.723,
 4.06,
 1.501,
 0.469,
 0.116,
 0.020,
 0.002]
. These steps are considered sufficient for studying the diffusion mappings for two reasons: \emph{(i)} images generated using these 10 steps closely match those generated with more steps, and \emph{(ii)} recent research~\cite{go2024addressing} demonstrates that the diffusion denoisers trained on similar noise variances exhibit analogous function mappings, implying that denoiser behavior at discrete variances represents their behavior at nearby variances.

After obtaining the linear models $\mathcal D_{\mathrm{L}}$, we evaluate their differences with the actual nonlinear denoisers $\mathcal D_{\mb \theta}$ with the score field approximation error, calculated using the expectation over the root mean square error (RMSE):
\begin{align} \label{Score Error}
    \text{Score-Difference}(t):= \mathbb{E}_{\mb x\sim p_{\text{data}}(\mb x),\mb \epsilon\sim \mathcal{N}(\mb 0;\sigma(t)^2\mb I)} \underbrace{\sqrt{ \frac{\| \mathcal D_{\mathrm{L}}(\mb x+\mb \epsilon;\sigma(t))-\mathcal D_{\mb \theta}(\mb x+\mb\epsilon;\sigma(t))\|_2^2}{d}}}_{\text{RMSE of a pair of randomly sampled $\mb x$ and $\mb \epsilon$}},
\end{align}
where $d$ represents the data dimension and the expectation is approximated with its empirical mean. While we present RMSE-based results in the main text, our findings remain consistent across alternative metrics, including NMSE, as detailed in~\Cref{Additional experiments}.

\begin{figure}[t]
    \centering
    \includegraphics[width=1\textwidth]{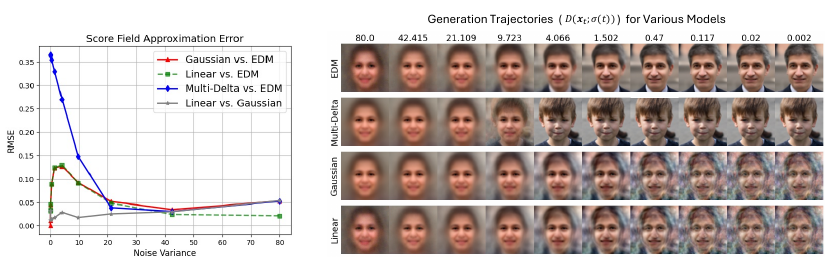}
    \vspace{-0.2in}
    \caption{\textbf{Score field approximation error and sampling Trajectory.} The left and right figures demonstrate the score field approximation error and the sampling trajectories $\mc D(\mb x_t;\sigma(t)$ of actual diffusion model (EDM), Multi-Delta model, linear model and Gaussian model respectively. Notice that the curve corresponding to the Gaussian model almost overlaps with that of the linear model, suggesting they share similar funciton mappings.}
    \label{fig: score error and sampling trajectory}
    \vspace{-0.2in}
\end{figure}

We perform linear distillation on well trained diffusion models operating in the generalization regime. For comprehensive analysis, we also compute the score approximation error between $\mathcal D_{\mb \theta}$ and: \emph{(i)} the optimal denoisers for the multi-delta distribution $\mathcal D_{\mathrm{M}}$ defined as \eqref{multi-delta denoiser}, and \emph{(ii)} the optimal denoisers for the multivariate Gaussian distribution $\mathcal D_{\mathrm{G}}$ defined as \eqref{Gaussian denoiser}. As shown in~\Cref{fig: score error and sampling trajectory}, our analysis reveals three distinct regimes:
\begin{itemize}[leftmargin=*]
    \item \emph{High-noise regime [20,80].} In this regime, only coarse image structures are generated (\Cref{fig: score error and sampling trajectory}(right)).
    Quantitatively, as shown in~\Cref{fig: score error and sampling trajectory}(left), the distilled linear model $\mathcal D_{\mathrm{L}}$ closely approximates its nonlinear counterpart $\mathcal D_{\mb \theta}$ with RMSE below 0.05. Both Gaussian score $\mathcal D_{\mathrm{G}}$ and multi-delta score $\mathcal D_{\mathrm{M}}$ also achieve comparable approximation accuracy.
    \item \emph{Low-noise regime [0.002,0.1].} In this regime, only subtle, imperceptible details are added to the generated images. Here, both $\mathcal D_{\mathrm{L}}$ and $\mathcal D_{\mathrm{G}}$ effectively approximate $\mathcal D_{\mb \theta}$ with RMSE below 0.05.

    \item \emph{Intermediate-noise regime [0.1,20]:} This crucial regime, where realistic image content is primarily generated, exhibits significant nonlinearity. While $\mathcal{D_{\mathrm{M}}}$ exhibits high approximation error due to rapid convergence to training samples—a memorization effect theoretically proved in~\cite{gu2023memorization}, both $\mathcal D_{\mathrm{L}}$ and $\mathcal{D_{\mathrm{G}}}$ maintain relatively lower approximation errors.
 
\end{itemize}

Qualitatively, as shown in~\Cref{fig: score error and sampling trajectory}(right), despite the relatively high score approximation error in the intermediate noise regime, the images generated with $\mc D_{\mathrm{L}}$ closely resemble those generated with $\mc D_{\mb \theta}$ in terms of the overall image structure and certain amount of fine details. This implies \emph{(i)} the underlying linear structure within the nonlinear diffusion models plays a pivotal role in their generalization capabilities and \emph{(ii)} such linear structure is effectively captured by our distilled linear models. In the next section, we will explore this linear structure by examining the linear models $\mc D_{\mathrm{L}}$. 
\vspace{-0.1in}
\subsection{Inductive Bias towards Learning the Gaussian Structures}
\label{inductive bias towards Gaussian}
\vspace{-0.1in}

\begin{wrapfigure}[16]{r}{0.4\textwidth}
  \vspace{-0.5cm}
  \begin{center}
    \includegraphics[width=0.4\textwidth]{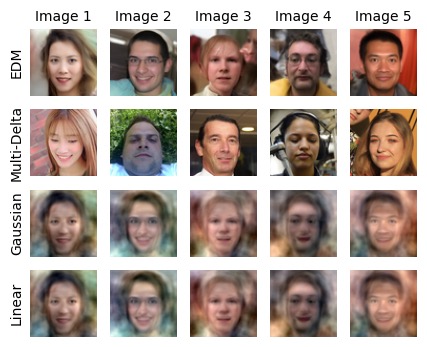}
  \end{center}
  \vspace{-0.3cm}
  \caption{\textbf{Images sampled from various Models.} The figure shows the samples generated using different models starting from the same initial noises.}
  \label{fig:linear v.s. Gaussian Samples}
\end{wrapfigure}

\begin{figure}[t]
    \vspace{-0.5cm}
    \centering
    \includegraphics{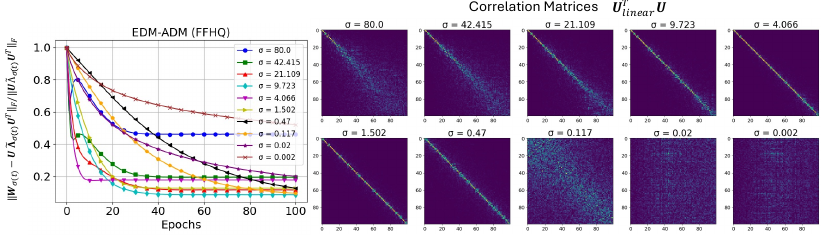}
    \caption{\textbf{Linear model shares similar function mapping with Gaussian model.} The left figure shows the difference between the linear weights and the Gaussian weights w.r.t. 100 training epochs of the linear distillation process for the 10 discrete noise levels. The right figure shows the correlation matrices between the first 100 singular vectors of the linear weights and Gaussian weights.}
    \vspace{-0.7cm}
    \label{fig:linear_vs_gaussian_2}
\end{figure}

Notably, the Gaussian denoisers $\mathcal{D_\mathrm{G}}$ exhibit behavior strikingly similar to the linear denoisers $\mc D_{\mathrm{L}}$. As illustrated in~\Cref{fig: score error and sampling trajectory}(left), they achieve nearly identical score approximation errors, particularly in the critical intermediate variance region. Furthermore, their sampling trajectories are remarkably similar (\Cref{fig: score error and sampling trajectory}(right)), producing nearly identical generated images that closely match those from the actual diffusion denoisers (\Cref{fig:linear v.s. Gaussian Samples}). These observations suggest that $\mc D_{\mathrm{L}}$ and $\mc D_{\mathrm{G}}$ share similar function mappings across various noise levels, leading us to hypothesize that the intrinsic linear structure underlying diffusion models corresponds to the Gaussian structure of the training data—specifically, its empirical mean and covariance.
We validate this hypothesis by empirically showing that $\mc D_{\mathrm{L}}$ is close to $\mc D_{\mathrm{G}}$ through the following three complementary experiments:

\noindent \textbullet\ \hspace{2pt}\emph{Similarity in weight matrices.} As illustrated in~\Cref{fig:linear_vs_gaussian_2}(left), $\mb W_{\sigma(t)}$ progressively converge towards $\mb U\Tilde{\Lambda}_{\sigma(t)}\mb U^T$ throughout the linear distillation process, achieving small normalized MSE (less than 0.2) for most of the noise levels. The less satisfactory convergence behavior at $\sigma(t)=80.0$ is due to inadequate training of the diffusion models at this particular noise level, which is minimally sampled during the training of actual diffusion models (see~\Cref{sec: Gaussian inductive bias as a general property} for more details).

\noindent \textbullet\ \hspace{2pt} \emph{Similarity in Score functions.}
Furthermore, \Cref{fig: score error and sampling trajectory}(left, gray line) demonstrates that $\mc D_{\mathrm{L}}$ and $\mc D_{\mathrm{G}}$ maintain small score differences (RMSE less than 0.05) across all noise levels, indicating that these denoisers exhibit similar function mappings throughout the diffusion process.

\noindent \textbullet\ \hspace{2pt} \emph{Similarity in principal components.} As shown in~\Cref{fig:linear_vs_gaussian_2}(right), for a wide noise range ($\sigma(t) \in [0.116, 80.0]$), the leading singular vectors of the linear weights $\mb{W}_{\sigma(t)}$ (denoted $\mb{U}_\text{\emph{Linear}}$) align well with $\mb{U}$, the singular vectors of the Gaussian weights.\footnote{For $\sigma(t)\in [0.116,80.0]$, the less well recovered singular vectors have singular values close to $0$, whereas those corresponding to high singular values are well recovered.} This implies that $\mb{U}$, representing the principal components of the training data, is effectively captured by the diffusion models. In the low-noise regime ($\sigma(t) \in [0.002, 0.116]$), however, $\mc{D}_{\mb{\theta}}$ approximates the identity mapping, leading to ambiguous singular vectors with minimal impact on image generation. Further analysis of $\mc{D}_{\mb{\theta}}$'s behavior in the low-noise regime is provided in \Cref{sec: identity behavior,Behaviors in low-noise regime}.

Since the optimization problem~\eqref{eq:linear distillation objective} is convex w.r.t. $\mb W_{\sigma(t)}$ and $\mb b_{\sigma(t)}$, the optimal solution $\mc D_{\mathrm{L}}$ represents the unique optimal linear approximation of $\mathcal{D}_{\mb \theta}$. Our analyses demonstrate that this optimal linear approximation closely aligns with $\mathcal{D}_{\mathrm{G}}$, leading to our central finding: diffusion models in the generalization regime exhibit an inductive bias (which we term as the Gaussian inductive bias) towards learning the Gaussian structure of training data. This manifests in two main ways: (\emph{i}) In the high-noise variance regime, well-trained diffusion models learn $\mathcal{D}_{\mb \theta}$ that closely approximate the linear Gaussian denoisers $\mathcal{D}_\mathrm{G}$; (\emph{ii}) As noise variance decreases, although $\mathcal{D}_{\mb \theta}$ diverges from $\mathcal{D}_\mathrm{G}$, $\mathcal{D}_\mathrm{G}$ remains nearly identical to the optimal linear approximation $\mathcal{D_\mathrm{L}}$, and images generated by $\mathcal{D_\mathrm{G}}$ retain structural similarity to those generated by $\mathcal{D}_{\mb \theta}$.

\begin{figure}
    \centering
    \vspace{-0.5in}
    \includegraphics[width=0.9\linewidth]{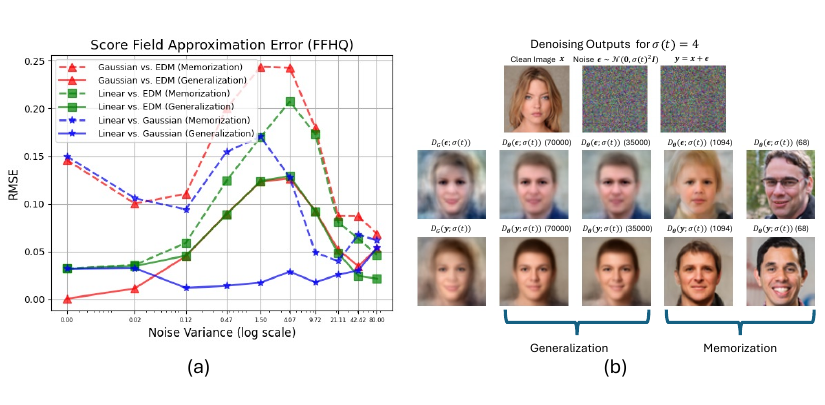}
    \caption{\textbf{Comparison between the diffusion denoisers in memorization and generalization regimes}. Figure(a) demonstrates that in the memorization regime (trained on small datasets of size 1094 and 68), $\mathcal D_{\mathrm{L}}$ significantly diverges from $\mathcal D_{\mathrm{G}}$, and both provide substantially poorer approximations of $\mc D_{\mb \theta}$ compared to the generalization regime (trained on larger datasets of size 35000 and 1094). Figure(b) qualitatively shows that the denoising outputs of $\mc D_{\mb \theta}$ closely match those of $\mathcal D_{\mathrm{G}}$ only in the generalization regime—a similarity that persists even when the denoisers process pure noise inputs.}
    \vspace{-0.5cm}
    \label{fig: uniqueness}
\end{figure}

Finally, we emphasize that the Gaussian inductive bias only emerges in the generalization regime. By contrast, in the memorization regime, \Cref{fig: uniqueness} shows that $\mc D_{\mathrm{L}}$ significantly diverges from $\mathcal{D_{\mathrm{G}}}$, and both $\mc D_{\mathrm{G}}$ and $\mc D_{\mathrm{L}}$ provide considerably poorer approximations of $\mathcal{D_{\mb \theta}}$ compared to the generalization regime.


\vspace{-0.1in}
\subsection{Theoretical Analysis} 
\label{theoretical analysis}
\vspace{-0.1in}
In this section, we demonstrate that imposing linear constraints on diffusion models while minimizing the denoising score matching objective~\eqref{Training Denoisers} leads to the emergence of Gaussian structure.

\begin{thm}
    Consider a diffusion denoiser parameterized as a single-layer linear network, defined as $\mathcal D(\mb x_t;\sigma(t))=\mb W_{\sigma(t)}\mb x_t+\mb b_{\sigma(t)}$, where $\mb W_{\sigma(t)} \in \mathbb R^{d \times d}$ is a linear weight matrix and  $\mb b_{\sigma(t)} \in \bb R^d $ is the bias vector. When the data distribution $p_\text{data}(\mb x)$ has finite mean $\mb \mu$ and bounded positive semidefinite covariance $\mb\Sigma$, the optimal solution to the score matching objective \eqref{Training Denoisers} is exactly the Gaussian denoiser defined in~\eqref{Gaussian denoiser}:
    \begin{align*}
        \mathcal D_{\mathrm{G}}(\mb x_t;\sigma(t)) \;=\; \mb U\Tilde{ \mb \Lambda}_{\sigma(t)}\mb U^T (\mb x_t-\mb \mu) + \mb \mu,
    \end{align*}
    with $\mb W_{\sigma(t)} = \mb U\Tilde{\mb \Lambda}_{\sigma(t)}\mb U^T$ and $\mb b_{\sigma(t)} =  \paren{ \mb I - \mb U\Tilde{\mb \Lambda}_{\sigma(t)}\mb U^T} \mb \mu $. 
\end{thm}

The detailed proof is postponed to~\Cref{detailed proof}. This optimal solution corresponds to the classical Wiener filter~\cite{STEPHANE2009535}, revealing that diffusion models naturally learn the Gaussian denoisers when constrained to linear architectures. To understand why highly nonlinear diffusion models operate near this linear regime, it is helpful to model the training data distribution as the multi-delta distribution $p(\mb x)=\frac{1}{N}\sum_{i=1}^N\delta(\mb x-\mb y_i)$, where $\{\mb y_1, \mb y_2,...,\mb y_N\}$ is the finite training images. Notice that this formulation better reflects practical scenarios where only a finite number of training samples are available rather than the ground truth data distribution. Importantly, it is proved in~\cite{wang2023hidden} that the optimal denoisers $\mathcal D_{\mathrm{M}}$ in this case is approximately equivalent to $\mc{D}_{\mathrm{G}}$ for high noise variance $\sigma (t)$ and query points far from the finite training data. This equivalence explains the strong similarity between $\mc{D}_{\mathrm{G}}$ and $D_{\mathrm{M}}$ in the high-noise variance regime, and consequently, why $\mathcal{D}_{\mb \theta}$ and $\mc{D}_{\mathrm{G}}$ exhibit high similarity in this regime---deep networks converge to the optimal denoisers for finite training datasets.

However, this equivalence between $\mc{D}_{\mathrm{G}}$ and $\mathcal{D}_{\mathrm{M}}$ breaks down at lower $\sigma(t)$ values. The denoising outputs of $\mathcal{D}_{\mathrm{M}}$ are convex combinations of training data points, weighted by a softmax function with temperature $\sigma(t)^2$. As $\sigma(t)^2$ decreases, this softmax function increasingly approximates an argmax function, effectively retrieving the training point $\mb y_i$ closest to the input $\mb x$. Learning this optimal solution requires not only sufficient model capacity to memorize the entire training dataset but also, as shown in~\cite{zeno2024minimum}, an exponentially large number of training samples. Due to these learning challenges, deep networks instead converge to local minima $\mathcal{D}_{\mb \theta}$ that, while differing from $\mathcal{D}_{\mathrm{M}}$, exhibit better generalization property. Our experiments reveal that these learned $\mathcal{D}_{\mb \theta}$ share similar function mappings with $\mc{D}_{\mathrm{G}}$. The precise mechanism driving diffusion models trained with gradient descent towards this particular solution remains an open question for future research.

Notably, modeling $p_\text{data}(\mb x)$ as a multi-delta distribution reveals a key insight: while unconstrained optimal denoisers~\eqref{multi-delta denoiser} perfectly capture the scores of the empirical distribution, they have no generalizability. In contrast, Gaussian denoisers, despite having higher score approximation errors due to the linear constraint, can generate novel images that closely match those produced by the actual diffusion models. This suggests that the generative power of diffusion models stems from the imperfect learning of the score functions of the empirical distribution.

\vspace{-0.1in}
\section{Conditions for the Emergence of Gaussian Structures and Generalizability}
\label{sec: conditon for the inductive bias}
\vspace{-0.1in}

In \Cref{sec: Inductive bias towards learning Gaussian}, we demonstrate that diffusion models exhibit an inductive bias towards learning denoisers that are close to the Gaussian denoisers. In this section, we investigate the conditions under which this bias manifests. Our findings reveal that this inductive bias is linked to model generalization and is governed by \emph{(i)} the model capacity relative to the dataset size and \emph{(ii)} the training duration. For additional results, including experiments on CIFAR-10 dataset, see \Cref{additional results sec 4}.
\vspace{-0.1in}
\subsection{Gaussian Structures Emerge when Model Capacity is Relatively Small}
\vspace{-0.1in}
First, we find that the Gaussian inductive bias and the generalization of diffusion models are heavily influenced by the relative size of the model capacity compared to the training dataset. In particular, we demonstrate that:
\begin{center}
    \emph{\textbf{Diffusion models learn the Gaussian structures when the model capacity is relatively small compared to the size of training dataset.}}
\end{center}
\begin{figure}[t]
    \centering
     \vspace{-0.1cm}
    \includegraphics[width=0.9\linewidth]{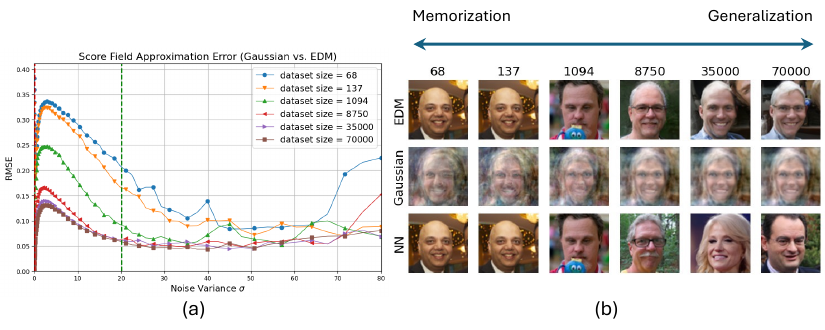}
    \caption{\textbf{Diffusion models learn the Gaussian structure when training dataset is large.} Models with a fixed scale (channel size 128) are trained across various dataset sizes. The left and right figures show the score difference and the generated images respectively. "NN" denotes the nearest neighbor in the training dataset to the images generated by the diffusion models.}
    \vspace{-0.1in}
    \label{fig:various_dataset_size}
\end{figure}
This argument is supported by the following two key observations:

\noindent \textbullet\ \hspace{2pt}\emph{\textbf{Increasing dataset size prompts the emergence of Gaussian structure at fixed model scale.}}
    We train diffusion models using the EDM configuration~\cite{karras2022elucidating} with a fixed channel size of 128 on datasets of varying sizes $[68, 137, 1094, 8750, 35000, 70000]$ until FID convergence. \Cref{fig:various_dataset_size}(left) demonstrates that the score approximation error between diffusion denoisers $\mc D_{\mb \theta}$ and Gaussian denoisers $\mc D_{\mathrm{G}}$ decreases as the training dataset size grows, particularly in the crucial intermediate noise variance regime ($\sigma(t)\in [0.116,20]$). This increasing similarity between $\mc D_{\mb \theta}$ and $\mc D_{\mathrm{G}}$ correlates with a transition in the models' behavior: from a memorization regime, where generated images are replicas of training samples, to a generalization regime, where novel images exhibiting Gaussian structure\footnote{We use the term "exhibiting Gaussian structure" to describe images that resemble those generated by Gaussian denoisers.} are produced, as shown in \Cref{fig:various_dataset_size}(b). This correlation underscores the critical role of Gaussian structure in the generalization capabilities of diffusion models.
    
\begin{figure}[t]
    \centering
    \vspace{-0.1in}
    \includegraphics[width=0.9\linewidth]{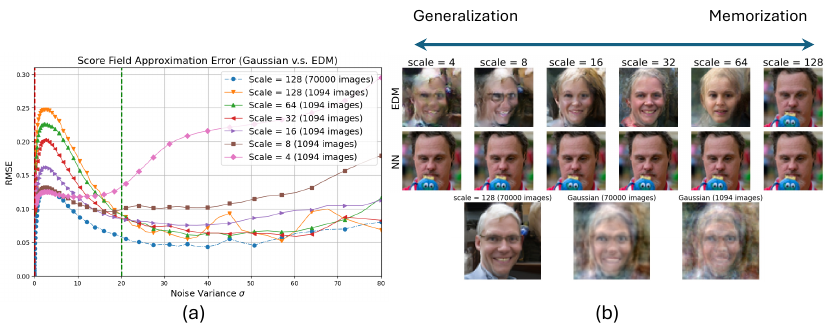}
    \caption{\textbf{Diffusion model learns the Gaussian structure when model scale is small.} Models with different scales are trained on a fixed training dataset of 1094 images. The left and right figures show the score difference and the generated images respectively.}
    \label{fig:various_model_scale}
    \vspace{-0.1in}
\end{figure}

\noindent \textbullet\ \hspace{2pt}\emph{\textbf{Decreasing model capacity promotes the emergence of Gaussian structure at fixed dataset sizes.}} Next, we investigate the impact of model scale by training diffusion models with varying channel sizes $[4, 8, 16, 32, 64, 128]$, corresponding to $[64\text{k}, 251\text{k}, 992\text{k}, 4\text{M}, 16\text{M}, 64\text{M}]$ parameters, on a fixed training dataset of 1094 images. \Cref{fig:various_model_scale}(left) shows that in the intermediate noise variance regime ($\sigma(t) \in[0.116,20]$), the discrepancy between $\mc D_{\mb \theta}$ and $\mc D_{\mathrm{G}}$ decreases with decreasing model scale, indicating that Gaussian structure emerges in low-capacity models. \Cref{fig:various_model_scale}(right) demonstrates that this trend corresponds to a transition from data memorization to the generation of images exhibiting Gaussian structure. Here we note that smaller models lead to larger discrepancy between $\mc D_{\mb \theta}$ and $\mc D_{\mathrm{G}}$ in the high-noise regime. This phenomenon arises because diffusion models employ a bell-shaped noise sampling distribution that prioritizes intermediate noise levels, resulting in insufficient training at high noise variances, especially when model capacity is limited (see more details in~\Cref{sec: behavior in high-noise reigme}).

These two experiments collectively suggest that the inductive bias of diffusion models is governed by the relative capacity of the model compared to the training dataset size.
\vspace{-0.1in}
\subsection{Overparameterized Models Learn Gaussian Structures before Memorization}
\vspace{-0.1in}

\begin{figure}[t]
    \centering
    \vspace{-0.2cm}
    \includegraphics[width=0.9\linewidth]{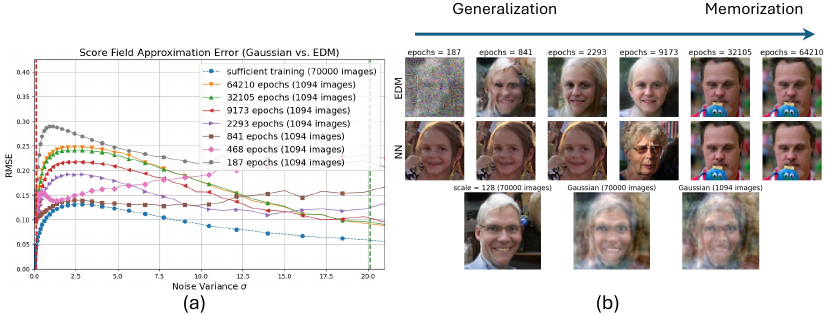}
    \caption{\textbf{Diffusion model learns the Gaussian structure in early training epochs.} Diffusion model with same scale (channel size 128) is trained using 1094 images. The left and right figures shows the score difference and the generated images respectively.}
    \vspace{-0.2cm}
    \label{fig:various_epochs}
\end{figure}

In the overparameterized regime, where model capacity significantly exceeds training dataset size, diffusion models eventually memorize the training data when trained to convergence. However, examining the learning progression reveals a key insight:
\begin{center}
    \emph{\textbf{Diffusion models learn the Gaussian structures with generalizability before they memorize.}
}
\end{center}
\Cref{fig:various_epochs}(a) demonstrates that during early training epochs (0-841), $\mc D_{\mb \theta}$ progressively converge to $\mc D_{\mathrm{G}}$ in the intermediate noise variance regime, indicating that the diffusion model is progressively learning the Gaussian structure in the initial stages of training. Notably. By epoch 841, the diffusion model generates images strongly resembling those produced by the Gaussian model, as shown in \Cref{fig:various_epochs}(b). However, continued training beyond this point increases the difference between $\mc D_{\mb \theta}$ and $\mc D_{\mathrm{G}}$ as the model transitions toward memorization. This observation suggests that early stopping could be an effective strategy for promoting generalization in overparameterized diffusion models.


\vspace{-0.1in}
\section{Connection between Strong Generalizability and Gaussian Structure}
\vspace{-0.1in}
\label{strong generalizability}

A recent study~\cite{kadkhodaie2023generalization} reveals an intriguing "strong generalization" phenomenon: diffusion models trained on large, non-overlapping image datasets generate nearly identical images from the same initial noise. While this phenomenon might be attributed to deep networks' inductive bias towards learning the "true" continuous distribution of photographic images, we propose an alternative explanation: rather than learning the complete distribution, deep networks may capture certain low-dimensional common structural features shared across these datasets and these features can be partially explained by the Gaussian structure. 

\begin{figure}[t]
    \centering
    \vspace{-0.1in}
    \includegraphics[width=\textwidth]{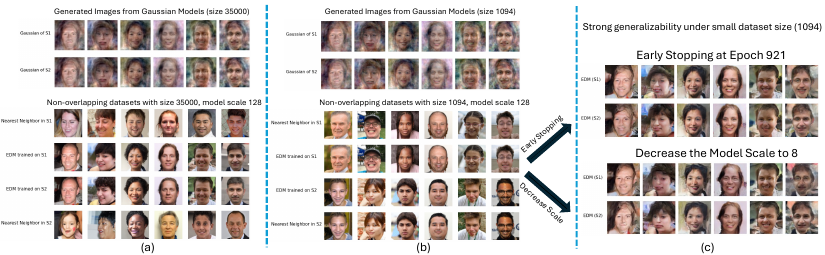}
    \caption{\textbf{Diffusion models in the strong generalization regime generate similar images as the Gaussian models.} Figure(a) Top: Generated images of Gausisan models; Bottom: Generated images of diffusion models, with model scale 128; S1 and S2 each has 35000 non-overlapping images. Figure(b) Top: Generated images of Gausisan model; Bottom: Generated images of diffusion models in the memorization regime, with model scale 128; S1 and S2 each has 1094 non-overlapping images. Figure(c): Early stopping and reducing model capacity help transition diffusion models from memorization to generalization.}
    \vspace{-0.5cm}
    \label{fig:strong gneralizability}
\end{figure}

To validate this hypothesis, we examine two diffusion models with channel size 128, trained on non-overlapping datasets S1 and S2 (35000 images each). \Cref{fig:strong gneralizability}(a) shows that images generated by these models (bottom) closely match those from their corresponding Gaussian models (top), highlighting the Gaussian structure's role in strong generalization.

Comparing \Cref{fig:strong gneralizability}(a)(top) and (b)(top), we observe that $\mc D_{\mathrm{G}}$ generates nearly identical images whether the Gaussian structure is calculated on a small dataset (1094 images) or a much larger one (35000 images). This similarity emerges because datasets of the same class can exhibit similar Gaussian structure (empirical covariance) with relatively few samples—just hundreds for FFHQ. Given the Gaussian structure's critical role in generalization, small datasets may already contain much of the information needed for generalization, contrasting previous assertions in~\cite{kadkhodaie2023generalization} that strong generalization requires training on datasets of substantial size (more than $10^5$ images). However, smaller datasets increase memorization risk, as shown in \Cref{fig:strong gneralizability}(b). To mitigate this, as discussed in \Cref{sec: conditon for the inductive bias}, we can either reduce model capacity or implement early stopping (\Cref{fig:strong gneralizability}(c)). Indeed, models trained on 1094 and 35000 images generate remarkably similar images, though the smaller dataset yields lower perceptual quality. This similarity further demonstrates that small datasets contain substantial generalization-relevant information closely tied to Gaussian structure.
Further discussion on the connections and differences between our work and~\cite{kadkhodaie2023generalization} are detailed in~\Cref{dicussion on GAHB}.

\vspace{-0.1in}
\section{Discussion}
\vspace{-0.1in}

 In this study, we empirically demonstrate that diffusion models in the generalization regime have the inductive bias towards learning diffusion denoisers that are close to the corresponding linear Gaussian denoisers. Although real-world image distributions are significantly different from Gaussian, our findings imply that diffusion models have the bias towards learning and utilizing low-dimensional data structures, such as the data covariance, for image generation. However, the underlying mechanism by which the nonlinear diffusion models, trained with gradient descent, exhibit such linearity remains unclear and warrants further investigation. 
 
 Moreover, the Gaussian structure only partially explains diffusion models' generalizability. While models exhibit increasing linearity as they transition from memorization to generalization, a substantial gap persists between the linear Gaussian denoisers and the actual nonlinear diffusion models, especially in the intermediate noise regime. As a result, images generated by Gaussian denoisers fall short in perceptual quality compared to those generated by the actual diffusion models especially for complex dataset such as CIFAR-10. This disparity highlights the critical role of nonlinearity in high-quality image generation, a topic we aim to investigate further in future research.

\section*{Data Availability Statement}
The code and instructions for reproducing the experiment results will be made available in the following link: \url{https://github.com/Morefre/Understanding-Generalizability-of-Diffusion-Models-Requires-Rethinking-the-Hidden-Gaussian-Structure}.
\section*{Acknowledgment}
We acknowledge funding support from NSF CAREER CCF-2143904, NSF CCF-2212066, NSF CCF-
2212326, NSF IIS 2312842, NSF IIS 2402950, ONR N00014-22-1-2529, a gift grant from KLA, an
Amazon AWS AI Award, and MICDE Catalyst Grant. We also acknowledge the computing support from NCSA Delta GPU~\cite{boerner2023access}. We thank Prof. Rongrong Wang (MSU) for fruitful discussions and valuable feedbacks.
\bibliographystyle{unsrt}
\bibliography{nips_main}


\clearpage
\appendix
\begin{center}
{\LARGE \bf Appendices}
\end{center}\vspace{-0.15in}
\par\noindent\rule{\textwidth}{1pt}
\tableofcontents

\section{Measuring the Linearity of Diffusion Denoisers}
\label{sec: linearity of diffusion denoisers}
In this section, we provide a detailed discussion on how to measure the linearity of diffusion model. For a diffusion denoiser, $\mc D_{\mb \theta}(\mb x;\sigma(t))$, to be considered approximately linear, it must fulfill the following conditions:

\begin{itemize}[leftmargin=*]
\item \emph{Additivity:} The function should satisfy $\mc D_{\mb \theta}(\mb x_1+\mb x_2;\sigma(t)) \approx \mc D_{\mb \theta}(\mb x_1;\sigma(t)) + \mc D_{\mb \theta}(\mb x_2;\sigma(t))$.
\item \emph{Homogeneity:} It should also adhere to $\mc D_{\mb \theta}(\alpha \mb x;\sigma(t)) \approx \alpha\mc D_{\mb \theta}(\mb x;\sigma(t))$.
\end{itemize}
To jointly assess these properties, we propose to measure the difference between $\mathcal D_{\mb \theta} (\alpha \mb x_1+\beta \mb x_2; \sigma (t))$ and $\alpha \mathcal D_{\mb \theta}(\mb x_1;\sigma(t))+\beta D(\mb x_2;\sigma(t))$.  While the linearity score is introduced as the cosine similarity between $\mathcal D_{\mb \theta} (\alpha \mb x_1+\beta \mb x_2; \sigma (t))$ and $\alpha \mathcal D_{\mb \theta}(\mb x_1;\sigma(t))+\beta D(\mb x_2;\sigma(t))$ in the main text:

\begin{align}
  \label{linear score cosine}
  \texttt{LS}(t) \;=\; \bb E_{ \mb x_1, \mb x_2 \sim p(\mb x;\sigma(t)) }\brac{ \abs{\innerprod{ \frac{\mathcal D_{\mb \theta} (\alpha \mb x_1+\beta \mb x_2; \sigma (t))}{ \norm{\mathcal D_{\mb \theta} (\alpha \mb x_1+\beta \mb x_2; \sigma (t))}{2} }  }{ \frac{ \alpha \mathcal D_{\mb \theta}(\mb x_1;\sigma(t))+\beta \mathcal D_{\mb \theta}(\mb x_1;\sigma(t)) }{ \norm{\alpha \mathcal D_{\mb \theta}(\mb x_1;\sigma(t))+\beta \mathcal D_{\mb \theta}(\mb x_1;\sigma(t)) }{2} }} } },
\end{align}
it can also be defined with the normalized mean square difference (NMSE):

\begin{align}
    \label{linear score mse}
    \bb E_{ \mb x_1, \mb x_2 \sim p(\mb x;\sigma(t)) }\frac{||\mathcal D_{\mb \theta} (\alpha \mb x_1+\beta \mb x_2; \sigma (t))-(\alpha \mathcal D_{\mb \theta}(\mb x_1;\sigma(t))+\beta \mathcal D_{\mb \theta}(\mb x_1;\sigma(t)))||_2}{||\mathcal D_{\mb \theta} (\alpha \mb x_1+\beta \mb x_2; \sigma (t))||_2},
\end{align}
where the expectation is approximated with its empirical mean over 100 randomly sampled pairs of $(\mb x_1,\mb x_2)$. In the next section, we will demonstrate the linearity score with both metrics.

\begin{figure}[b]
    \centering
    \includegraphics{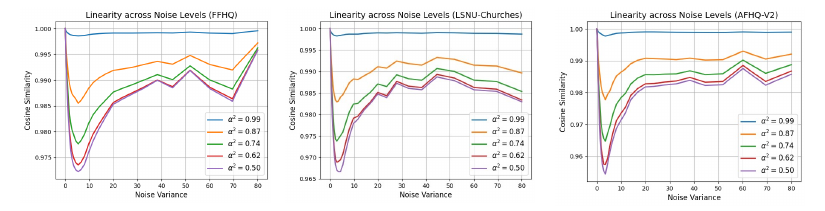}
    \caption{\textbf{Linearity scores for varying $\alpha$ and $\beta$.} The diffusion models are trained with the edm-ve configuration~\cite{karras2022elucidating}, which ensures the models are in the generalization regime.}
    \label{fig:linearity alphas}
\end{figure}

Since the diffusion denoisers are trained solely on inputs $\mb x\sim p(\mb x;\sigma(t))$, their behaviors on out-of-distribution inputs can be quite irregular. 
To produce a denoised output with meaningful image structure, it is critical that the noise component in the input $\mb x$ matches the correct variance $\sigma(t)^2$. Therefore, our analysis of linearity is restricted to in-distribution inputs $\mb x_1$ and $\mb x_2$, which are randomly sampled images with additive Gaussian noises calibrated to noise variance $\sigma(t)^2$. We also need to ensure that the values of $\alpha$ and $\beta$ are chosen such that $\alpha^2 + \beta^2 = 1$, maintaining the correct variance for the noise term in the combined input $\alpha \mb x_1 + \beta \mb x_2$. We present the linearity scores, calculated with varying values of $\alpha$ and $\beta$, for diffusion models trained on diverse datasets in \Cref{fig:linearity alphas}. These models are trained with the EDM-VE configuration proposed in~\cite{karras2022elucidating}, which ensures the resulting models are in the generalization regime. Typically, setting \(\alpha = \beta = 1/\sqrt{2}\) yields the lowest linearity score; however, even in this scenario, the cosine similarity remains impressively high, exceeding 0.96. This high value underscores the presence of significant linearity within diffusion denoisers. 

\begin{figure}[t]
    \centering
    \includegraphics{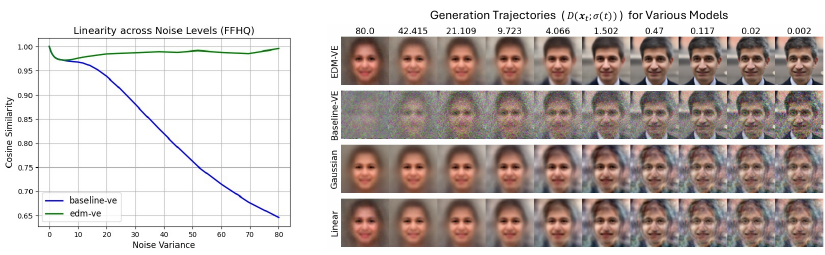}
    \caption{\textbf{Linearity scores and sampling trajectory.} The left and right figures demonstrate the linearity scores and the sampling trajectories $\mc D(\mb x_t;\sigma(t)$ of actual diffusion model (EDM-VE and Baseline-VE), Multi Delta model, linear model, and Gaussian model respectively. }
    \label{fig:baselin-ve vs. edm-ve}
\end{figure}

We would like to emphasize that for linearity to manifest in diffusion denoisers, it is crucial that they are well-trained, achieving a low denoising score matching loss as indicated in~\eqref{Training Denoisers}. As shown in~\Cref{fig:baselin-ve vs. edm-ve}, the linearity notably reduces in a less well trained diffusion model (Baseline-VE) comapred to its well-trained counterpart (EDM-VE). Although both models utilize the same 'VE' network architecture $\mc F_{\mb \theta}(\mb x;\sigma(t))$~\cite{ho2020denoising}, they differ in how the diffusion denoisers are parameterized:
\begin{align}
    \mc D_{\mb \theta}(\mb x;\sigma(t)):= c_\text{skip}(\sigma(t))\mb x+c_\text{out}(\mc F_{\mb \theta}(\mb x;\sigma(t))),
    \label{denoiser parameterization}
\end{align}
where $c_\text{skip}$ is the skip connection and $c_\text{out}$ modulate the scale of the network output. With carefully tailored $c_\text{skip}$ and $c_\text{out}$, the EDM-VE configuration achieves a lower score matching loss compared to Baseline-VE, resulting in samples with higher quality as illustrated in~\Cref{fig:baselin-ve vs. edm-ve}(right).

\begin{figure}[t]
    \centering
    \includegraphics[width=1\linewidth]{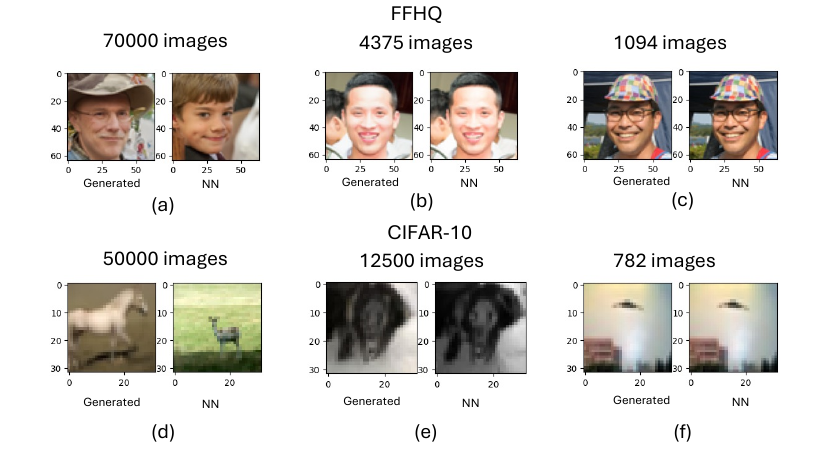}
    \caption{\textbf{Memorization and generalization regimes of diffusion models.} Figures(a) to (c) show the images generated by diffusion models trained on 70000, 4375, 1094 FFHQ images and their corresponding nearest neighbors in the training dataset respectively. Figures(d) to (f) show the images generated by diffusion models trained on 50000, 12500, 782 CIFAR-10 images and their corresponding nearest neighbors in the training dataset respectively. Notice that when the training dataset size is small, diffusion model can only generate images in the training dataset.}
    \label{generalization and memorization qualitative}
\end{figure}
\section{Emerging Linearity of Diffusion Models}
\label{emerging linearity from memorization to generalization}
In this section we provide a detailed discussion on the observation that diffusion models exhibit increasing linearity as they transition from memorization to generalization, which is briefly described in~\Cref{linear characteristic}. 
\subsection{Generalization and Memorization Regimes of Diffusion Models}
As shown in~\Cref{generalization and memorization qualitative}, as the training dataset size increases, diffusion models transition from the memorization regime—where they can only replicate its training images---to the generalization regime, where the they produce high-quality, novel images. 
To measure the generalization capabilities of diffusion models, it is crucial to assess their ability to generate images that are not mere replications of the training dataset. This can be quantitatively evaluated by generating a large set of images from the diffusion model and measuring the average difference between these generated images and their nearest neighbors in the training set. Specifically, let $\{\mb x_1, \mb x_2, ..., \mb x_k\}$ represent $k$ randomly sampled images from the diffusion models (we choose $k=100$ in our experiments), and let $Y:=\{\mb y_1, \mb y_2, ..., \mb y_N\}$ denote the training dataset consisting of $N$ images. We define the generalization score as follows:
\begin{align}
    \label{generalization score}
    \text{GL Score} := \frac{1}{k}\sum_{i=1}^k\frac{||\mb x_i-\text{NN}_Y(\mb x_i)||_2}{||\mb x_i||_2}
\end{align}
where $\text{NN}_Y (\mb x_i)$ represents the nearest neighbor of the sample $\mb x_k$ in the training dataset $Y$, determined by the Euclidean distance on a per-pixel basis. Empirically, a GL score exceeding 0.6 indicates that the diffusion models are effectively generalizing beyond the training dataset.
\begin{figure}[t]
    \centering
    \includegraphics[width=1\linewidth]{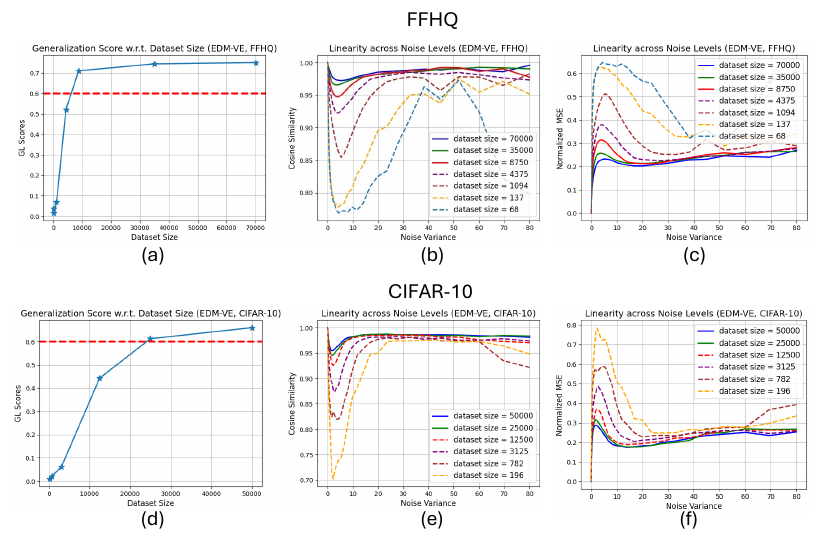}
    \caption{\textbf{Diffusion model exhibit increasing linearity as they transition from memorization to generalization.} Figure(a) and (d) demonstrate that for both FFHQ and CIFAR-10 datasets, the generalization score increases with the training dataset size, indicating progressive model generalization. Figure(b), (c), (e), and (f) show that this transition towards generalization is accompanied by increasing denoiser linearity. Specifically, Figure(b) and (e) display linearity scores calculated using cosine similarity~\eqref{linear score cosine}, while Figure(c) and (f) show scores computed using NMSE~\eqref{linear score mse}. Both metrics reveal consistent trends.
    }
    \label{fig:emerging linearity full}
\end{figure}
\subsection{Diffusion Models Exhibit Linearity in the Generalization Regime}
As demonstrated in~\Cref{fig:emerging linearity full}(a) and (d), diffusion models transition from the memorization regime to the generalization regime as the training dataset size increases. Concurrently, as depicted in~\Cref{fig:emerging linearity full}(b), (c), (e) and (f), the corresponding diffusion denoisers exhibit increasingly linearity. This phenomenon persists across diverse datasets datasets including FFHQ~\cite{karras2019style}, AFHQ~\cite{choi2020stargan} and LSUN-Churches~\cite{yu2015lsun}, as well as various model architectures including EDM-VE~\cite{songscore}, EDM-VP~\cite{ho2020denoising} and EDM-ADM~\cite{dhariwal2021diffusion}. This emerging linearity implies that the hidden linear structure plays an important role in the generalizability of diffusion model.

\section{Linear Distillation}
\label{sec: linear distillation}
As discussed in~\Cref{linear characteristic}, we propose to study the hidden linearity observed in diffusion denosiers with linear distillation. Specifically, for a given diffusion denoiser $\mc D_{\mb \theta}(\mb x;\sigma(t))$, we aim to approximate it with a linear function (with a bias term for more expressibility):
\begin{align*}
    \mathcal D_{\mathrm{L}}(\mb x;\sigma(t)) := \mb W_{\sigma(t)} \mb x+\mb b_{\sigma(t)}\;\approx\; \mathcal D_{\mb \theta} (\mb x;\sigma(t)),
\end{align*}
for $\mb x\sim p(\mb x;\sigma(t))$. Notice that for three dimensional images with size $(c,h,w)$, $\mb x\in \mathbb{R}^d$ represents their vectorized version, where $d=c\times w\times h$. Let
\begin{align*}
    \mathcal L(\bm W, \bm b) = \frac{1}{n} \sum_{i=1}^n \left\| \mb W_{\sigma(t)}\{k-1\} (\mb x_i + \mb \epsilon_i) + \mb b_{\sigma(t)}\{k-1\} - \mc D_{\mb\theta}(\mb x_i+\mb\epsilon_i; \sigma(t)) \right\|_2^2
\end{align*}

We train $10$ independent linear models for each of the selected noise variance level $\sigma(t)$ with the procedure summarized in~\Cref{alg: linear distillation}:
\begin{algorithm*}[h]
\caption{Linear Distillation} \label{alg: linear distillation}
\begin{algorithmic}
    \Require 
    \Statex (i) the targeted diffusion denoiser $\mc D_{\mb\theta}(\cdot;\sigma(t))$,
    \Statex (ii) weights $\mb W_{\sigma(t)}$ and biases $\mb b_{\sigma(t)}$, both initialized to zero,
    \Statex (iii) gradient step size $\eta$,
    \Statex (iv) number of training iterations $K$,
    \Statex (v) training batch size $n$,
    \Statex (vi) image dataset $S$.
    \For{$k = 1$ to $K$}
        \State Randomly sample a batch of training images $\{\mb x_1, \mb x_2, \ldots, \mb x_n\}$ from $S$.
        \State Randomly sample a batch of noises $\{\mb \epsilon_1, \mb \epsilon_2, \ldots, \mb\epsilon_n\}$ from $\mathcal{N}(\mb 0,\sigma(t)\mb I)$.
        \State Update $\mb W_{\sigma(t)}$ and $\mb b_{\sigma(t)}$ with gradient descent:
        \Statex $\mb W_{\sigma(t)}\{k\} = \mb W_{\sigma(t)}\{k-1\} - \eta \nabla_{\mb W_{\sigma(t)}\{k-1\}} \mathcal L(\bm W, \bm b)$
        \Statex $\mb b_{\sigma(t)}\{k\} = \mb b_{\sigma(t)}\{k-1\} - \eta \nabla_{\mb b_{\sigma(t)}\{k-1\}} \mathcal L(\bm W, \bm b)$
    \EndFor
    \State \textbf{Return} $\mb W_{\sigma(t)}\{K\}, \mb b_{\sigma(t)}\{K\}$
\end{algorithmic}
\end{algorithm*}

\begin{figure}[t]
    \centering
    \includegraphics[width=1\linewidth]{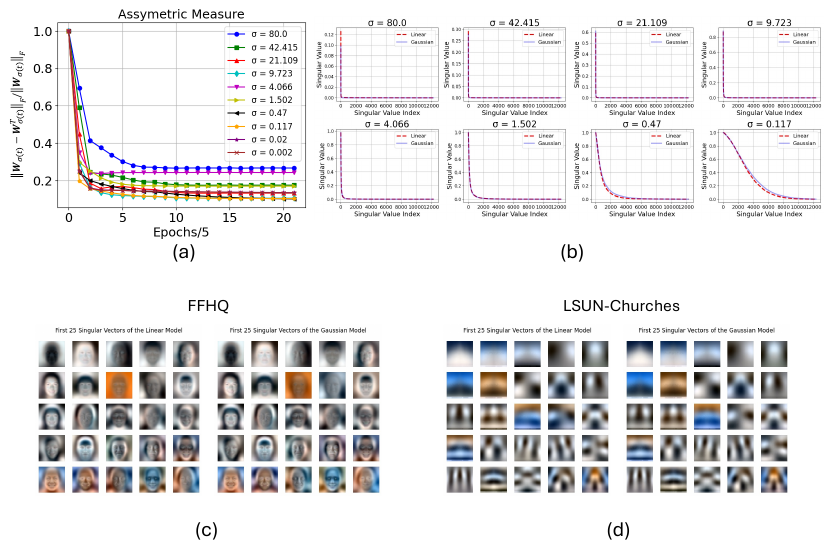}
    \caption{\textbf{Additional linear distillation results.} Figure(a) demonstrates the gradual symmetrization of linear weights during the distillation process. Figure(b) shows that at convergence, the singular values of the linear weights closely match those of the Gaussian weights. Figure(c) and Figure(d) display the leading singular vectors of both linear and Gaussian weights at $\sigma(t)=4$ for FFHQ and LSUN-Churches datasets, respectively, revealing a strong correlation.}
    \label{fig:additional linear distillation results}
\end{figure}

In practice, the gradients on $\mb W_{\sigma(t)}$ and $\mb b_{\sigma(t)}$ are obtained through automatic differentiation. Additionally, we employ the Adam optimizer~\cite{kingma2014adam} for updates. Additional linear distillation results are provided in~\Cref{fig:additional linear distillation results}.
\section{Diffusion Models in Low-noise Regime are Approximately Linear Mapping}
\label{sec: identity behavior}
\begin{figure}[t]
    \centering
    \includegraphics{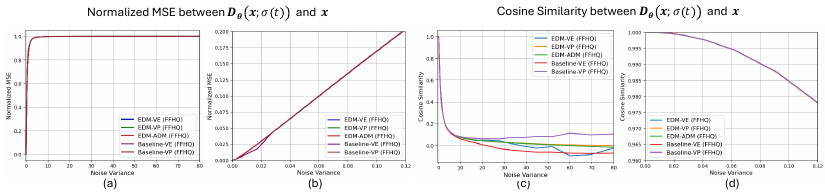}
    \caption{\textbf{Difference between $\mc D_{\mb \theta}(\mb x;\sigma(t))$ and $\mb x$ for various noise variance levels.} Figures(a) and (c) show the differences between $\mc D_{\mb \theta}(\mb x;\sigma(t))$ and $\mb x$ across $\sigma(t) \in [0.002, 80]$, measured by normalized MSE and cosine similarity, respectively. Figures(b) and (d) provide zoomed-in views of (a) and (c). The diffusion models were trained on the FFHQ dataset. Notice that the difference between $\mc D_{\mb \theta}(\mb x;\sigma(t))$ and $\mb x$ quickly converges to near zero in the low noise variance regime. The trend is consistent for various model architectures.}
    \label{fig:Identity Study.}
\end{figure}

It should be noted that the low score difference between $\mc D_{\mathrm{G}}$ and $\mc D_{\mb\theta}$ within the low-noise regime ($\sigma(t)\in [0.002,0.116]$) does not imply the diffusion denoisers capture the Gaussian structure, instead, the similarity arises since both of them are converging to the identity mapping as $\sigma (t)$ decreases. As shown in~\Cref{fig:Identity Study.}, within this regime, the differences between the noisy input $\mb x$ and their corresponding denoised outputs $\mc D_{\mb\theta}(\mb x;\sigma(t))$ quickly approach 0. This indicates that the learned denoisers $\mc D_{\mb\theta}$ progressively converge to the identity function. Additionally, from~\eqref{Gaussian denoiser}, it is evident that the difference between the Gaussian weights and the identity matrix diminishes as $\sigma(t)$ decreases, which explains why $\mc D_{\mathrm{G}}$ can well approximate $\mc D_{\mb\theta}$ in the low noise variance regime. 

We hypothesize that $\mc D_{\mb\theta}$ learns the identity function because of the following two reasons: 

\emph{(i)} within the low-noise regime, since the added noise is negligible compared to the clean image, the identity function already achieves a small denoising error, thus serving as a shortcut which is exploited by the deep network.

\emph{(ii)} As discussed in~\Cref{sec: linearity of diffusion denoisers}, diffusion models are typically parameterized as follows:
\begin{align*}
    \mc D_{\mb \theta}(\mb x;\sigma(t)):= c_\text{skip}(\sigma(t))\mb x+c_\text{out}(\mc F_{\mb \theta}(\mb x;\sigma(t))),
\end{align*}
where $\mc F_{\mb \theta}$ represents the deep network, and $c_\text{skip}(\sigma(t))$ and $c_\text{out}(\sigma(t))$ are adaptive parameters for the skip connection and output scaling, respectively, which adjust according to the noise variance levels. For canonical works on diffusion models~\cite{ho2020denoising, songscore, karras2022elucidating, dhariwal2021diffusion}, as $\sigma(t)$ approaches zero, $c_\text{skip}$ and $c_\text{out}$ converge to $1$ and $0$ respectively. Consequently, at low variance levels, the function forms of diffusion denoisers are approximatly identity mapping: $\mc D_{\mb \theta}(\mb x;\sigma(t))\approx \mb x$.

This convergence to identity mapping has several implications. First, the weights $\mb W_{\sigma(t)}$ of the distilled linear models $\mc D_\mathrm{L}$ approach the identity matrix at low variances, leading to ambiguous singular vectors. This explains the poor recovery of singular vectors for $\sigma(t)\in [0.002,0.116]$ shown in \Cref{fig:linear_vs_gaussian_2}. Second, the presence of the bias term in~\eqref{linear model with bias} makes it challenging for our linear model to learn the identity function, resulting in large errors at $\sigma(t)=0.002$ as shown in \Cref{fig:linear_vs_gaussian_2}(a).

Finally, from~\eqref{1st order update 3}, we observe that when $\mc D_{\mb \theta}$ acts as an identity mapping, $\mb x_{i+1}$ remains unchanged from $\mb x_i$. This implies that sampling steps in low-variance regions minimally affect the generated image content, as confirmed in \Cref{fig: score error and sampling trajectory}, where image content shows negligible variation during these steps.
\section{Theoretical Analysis}
\label{detailed proof}
\subsection{Proof of Theorem 1}
In this section, we give the proof of Theorem 1 (\Cref{theoretical analysis}). Our theorem is based on the following two assumptions:
\begin{assum}
Suppose that the diffusion denoisers are parameterized as single-layer linear networks, defined as $\mathcal D(\mb x;\sigma(t))=\mb W_{\sigma(t)}\mb x+\mb b_{\sigma(t)}$, where $\mb W_{\sigma(t)} \in \mathbb R^{d \times d}$ is the linear weight and  $\mb b_{\sigma(t)} \in \bb R^d $ is the bias.
\end{assum}

\begin{assum}
The data distribution $p_\text{data}(\mb x)$ has finite mean $\mb\mu$ and bounded positive semidefinite covariance $\mb \Sigma$
\end{assum}
\paragraph{Theorem 1.} \emph{Under Assumption 1 and Assumption 2, the optimal solution to the denoising score matching objective~\eqref{Training Denoisers} is exactly the Gaussian denoiser: $\mc D_{\mathrm{G}}(\mb x,\sigma(t))=\mb \mu+\mb U\Tilde{\mb\Lambda}_{\sigma(t)}\mb U^T (\mb x-\mb \mu)$,
where $\mb \Sigma=\mb U\mb\Lambda\mb U^T$ represents the SVD of the covariance matrix, with singular values $\lambda_{\{k=1,...,d\}}$ and $\Tilde{\mb\Lambda}_{\sigma(t)}=\diag[\frac{\lambda_k}{\lambda_k+\sigma(t)^2}]$. Furthermore, this optimal solution can be obtained via gradient descent with a proper learning rate.}

To prove \textbf{Theorem 1}, we first show that the Gaussian denoiser is the optimal solution to the denoising score matching objective under the linear network constraint. Then we will show that such optimal solution can be obtained via gradient descent with a proper learning rate.
\paragraph{The Global Optimal Solution.}Under the constraint that the diffusion denoiser is restricted to a single-layer linear network with bias:
    \begin{align}
        \mc D(\mb x;\sigma(t))=\mb W_{\sigma (t)}\mb x+\mb b_{\sigma(t)},
    \end{align}
We get the following optimizaiton problem from~\Cref{Training Denoisers}:
\begin{align}
    \label{eq: constrained optimization form}
    \mb W^\star, \mb b^\star = \argmin_{\mb W,\mb b}\mathcal{L}(\mb W,\mb b;\sigma(t)):=\mathbb{E}_{\mb x\sim p_\text{data}}\mathbb{E}_{\mb \epsilon\sim \mathcal{N}(\mb 0,\sigma(t)^2\mb I)} ||\mb W(\mb x+\mb \epsilon)+\mb b-\mb x||_2^2,
\end{align}
where we omit the footnote $\sigma(t)$ in $\mb W_{\sigma (t)}$ and $\mb b_{\sigma (t)}$ for simplicity. Since expectation preserves convexity, the optimization problem~\Cref{eq: constrained optimization form} is a convex optimization problem. To find the global optimum, we first eliminate $\mb b$ by requiring the partial derivative $\nabla_{\mb b}\mathcal{L}(\mb W,\mb b;\sigma(t))$ to be $\mb 0$. Since
\begin{align}
    \nabla_{\mb b}\mathcal{L}(\mb W,\mb b;\sigma(t))
    & = 2*\mathbb{E}_{\mb x\sim p_\text{data}}\mathbb{E}_{\mb \epsilon\sim \mathcal{N}(\mb 0,\sigma(t)^2\mb I)}((\mb W-\mb I)\mb x+\mb W\mb \epsilon+\mb b)\\
    &=  2*\mathbb{E}_{\mb x\sim p_\text{data}}((\mb W-\mb I)\mb x+\mb b)\\
    &= 2*((\mb W-\mb I)\mb \mu+\mb b),
\end{align}
we have
\begin{align}
    \label{eq:expression for b unconstrained}
    \mb b^\star = (\mb I-\mb W^*)\mb \mu.
\end{align}
Utilizing the expression for $\mb b$, we get the following equivalent form of the optimization problem:
\begin{align}
    \mb W^\star = \argmin_{\mb W}\mathcal{L}(\mb W;\sigma(t)):=2*\mathbb{E}_{\mb x\sim p_\text{data}}\mathbb{E}_{\mb \epsilon\sim \mathcal{N}(\mb 0,\sigma(t)^2\mb I)}||\mb W(\mb x-\mb \mu +\mb\epsilon)-(\mb x-\mb \mu)||_2^2.
\end{align}
The derivative $\nabla_{\mb W}\mathcal{L}(\mb W;\sigma(t))$ is:
\begin{align}
    \nabla_{\mb W}\mathcal{L}(\mb W;\sigma(t))&=2*\mathbb{E}_{\mb x}\mathbb{E}_{\mb \epsilon}(\mb W(\mb x-\mb \mu+\mb \epsilon)(\mb x-\mb \mu+\mb \epsilon)^T-(\mb x-\mb \mu)(\mb x-\mb \mu+\mb\epsilon)^T)\\
    &= 2*\mathbb{E}_{\mb x}((\mb W-\mb I)(\mb x-\mb \mu)(\mb x-\mb \mu)^T+\sigma(t)^2\mb W)\\
    &= 2*\mb W(\mb \Sigma+\sigma(t)^2\mb I)-2*\mb \Sigma.
 \end{align}                                          
 Suppose $\mb \Sigma=\mb U \mb\Lambda\mb U^T$ is the SVD of the empirical covariance matrix, with singular values $\lambda_{\{k=1,...,n\}}$, by setting $\nabla_{\mb W}\mathcal{L}(\mb W;\sigma(t))$ to $\mb 0$, we get the optimal solution:
\begin{align}
    \mb W^\star &= \mb U\mb\Lambda\mb U^T\mb U(\mb\Lambda+\sigma(t)^2\mb I)^{-1}\mb U^T\\
    &= \mb U \Tilde{\mb\Lambda}_{\sigma(t)} \mb U^T,
\end{align}
where $\Tilde{\mb\Lambda}_{\sigma(t)}[i,i]=\frac{\lambda_i}{\lambda_i+\sigma(t)^2}$ and $\lambda_i=\mb\Lambda[i,i]$. Substitute $\mb W^\star$ back to~\Cref{eq:expression for b unconstrained}, we have:
 \begin{align}
     \mb b^\star = (\mb I-\mb U\Tilde{\mb\Lambda}_{\sigma(t)}\mb U^T)\mb \mu.
 \end{align}
Notice that the expression for $\mb W^\star$ and $\mb b^\star$ is exactly the Gaussian denoiser. Next, we will show this optimal solution can be achieved with gradient descent.

\paragraph{Gradient Descent Recovers the Optimal Solution.}
 Consider minimizing the population loss:
 \begin{align}
 \label{eq: population risk}
     \mathcal{L}(\mb W,\mb b;\sigma(t)):=\mathbb{E}_{\mb x\sim p_{\text{data}}}\mathbb{E}_{\mb \epsilon\sim \mathcal{N}(\mb 0,\sigma(t)^2\mb I)}||\mb W(\mb x+\mb\epsilon)+\mb b-\mb x||_2^2.
 \end{align}
 Define $\tilde{\mb{W}} := \begin{bmatrix} \mb{W} & \mb{b} \end{bmatrix}$, $\tilde{\mb x}:=\begin{bmatrix} \mb x \\1\end{bmatrix}$ and $\Tilde{\mb \epsilon}=\begin{bmatrix} \mb\epsilon \\0\end{bmatrix}$, then we can rewrite~\Cref{eq: population risk} as:
 \begin{align}
 \label{eq: population risk_v2}
     \mathcal{L}(\tilde{\mb W};\sigma(t)):=\mathbb{E}_{\mb x\sim p_{\text{data}}}\mathbb{E}_{\mb \epsilon\sim \mathcal{N}(\mb 0,\sigma(t)^2\mb I)}||\tilde{\mb W}(\tilde{\mb x}+\Tilde{\mb\epsilon})-\mb x||_2^2.
 \end{align}
 We can compute the gradient in terms of $\tilde{\mb W}$ as:
 \begin{align}
     \nabla\mathcal{L}(\tilde{\mb W})&=2*\mathbb{E}_{\mb x, \mb \epsilon}(\Tilde{\mb W}(\Tilde{\mb x}+\Tilde{\mb \epsilon})(\Tilde{\mb x}+\Tilde{\mb \epsilon})^T-\mb x(\tilde{\mb x}+\tilde{\mb \epsilon})^T)\\
     &=2*\mathbb{E}_{\mb x, \mb \epsilon}(\tilde{\mb W}(\tilde{\mb x}\tilde{\mb x}^T+\tilde{\mb x}\tilde{\mb \epsilon}^T+\tilde{\mb \epsilon}\tilde{\mb x}^T+\tilde{\mb \epsilon}\tilde{\mb \epsilon}^T)-\mb x\tilde{\mb x}^T-\mb x\tilde{\mb \epsilon}^T).
 \end{align}
 Since $\mathbb{E}_{\mb \epsilon}(\tilde{\mb \epsilon})=\mb 0$ and $\mathbb{E}_{\mb \epsilon}(\tilde{\mb \epsilon}\tilde{\mb \epsilon}^T)=\begin{bmatrix}
     \sigma(t)^2\mb I_{d\times d} & \mb 0_{d\times 1} \\
     \mb 0_{1\times d} & 0
 \end{bmatrix}$, we have:
 \begin{align}
     \nabla\mathcal{L}(\tilde{\mb W})&=2*\mathbb{E}_{\mb x}(\tilde{\mb W}(\tilde{\mb x}\tilde{\mb x}^T+\begin{bmatrix}
     \sigma(t)^2\mb I_{d\times d} & \mb 0_{d\times 1} \\
     \mb 0_{1\times d} & 0
 \end{bmatrix})-\mb x\Tilde{\mb x}^T).
 \end{align}
 Since $\mathbb{E}(\tilde{\mb x}\tilde{\mb x}^T)=\begin{bmatrix}
     \mathbb{E}(\mb x \mb x^T) & \mathbb{E}(\mb x)\\
     \mathbb{E}(\mb x^T) & 1
 \end{bmatrix}$, we have:
 \begin{align}
     \nabla\mathcal{L}(\tilde{\mb W})=2\tilde{\mb W}\begin{bmatrix}
         \mathbb{E}_{\mb x}(\mb x\mb x^T)+\sigma(t)^2\mb I &\mb \mu \\
         \mb \mu^T & 1
     \end{bmatrix}-2\begin{bmatrix}
         \mathbb{E}_{\mb x}(\mb x^T\mb x)&\mb \mu
     \end{bmatrix}.
 \end{align}
 With learning rate $\eta$, we can write the update rule as:
 \begin{align}
     \tilde{\mb W}(t+1) &=  \tilde{\mb W}(t)(1-2\eta\begin{bmatrix}
         \mathbb{E}_{\mb x}(\mb x\mb x^T)+\sigma(t)^2\mb I &\mb \mu \\
         \mb \mu^T & 1
     \end{bmatrix})+2\eta\begin{bmatrix}
         \mathbb{E}_{\mb x}(\mb x^T\mb x)&\mb \mu
     \end{bmatrix}\\
     &= \tilde{\mb W}(t)(1-2\eta \mb A)+2\eta\begin{bmatrix}
         \mathbb{E}_{\mb x}(\mb x^T\mb x)&\mb \mu
     \end{bmatrix},
 \end{align}
 where we define $\mb A:=\mb I-2\eta\begin{bmatrix}
         \mathbb{E}_{\mb x}(\mb x\mb x^T)+\sigma(t)^2\mb I &\mb \mu \\
         \mb \mu^T & 1
     \end{bmatrix}$
for simplicity. By recursively expanding the expression for $\tilde{\mb W}$, we have:
\begin{align}
\tilde{\mb W}(t+1)=\tilde{\mb W}(0)\mb A^{t+1}+2\eta \begin{bmatrix}
         \mathbb{E}_{\mb x}(\mb x^T\mb x)&\mb \mu
     \end{bmatrix}\sum_{i=0}^{t}\mb A^i.   
\end{align}
Notice that there exists a $\eta$, such that every eigen value of $\mb A$ is smaller than 1 and greater than 0. In this case, $\mb A^{t+1}\rightarrow \mb 0$ as $t\rightarrow \infty$. Similarly, by the property of matrix geometric series, we have $\sum_{i=0}^t\mb A^i\rightarrow (\mb I-\mb A)^{-1}$. Therefore we have:
\begin{align}
    \tilde{\mb W}&\rightarrow \begin{bmatrix}
         \mathbb{E}_{\mb x}(\mb x^T\mb x)&\mb \mu
     \end{bmatrix} 
     \begin{bmatrix}
         \mathbb{E}_{\mb x}(\mb x\mb x^T)+\sigma(t)^2\mb I &\mb \mu \\
         \mb u^T & 1
     \end{bmatrix}^{-1}\\
     &= \begin{bmatrix}
         \mathbb{E}_{\mb x}(\mb x^T\mb x)&\mb \mu
     \end{bmatrix} 
     \begin{bmatrix}
         \mb B &\mb \mu \\
         \mb \mu^T & 1
     \end{bmatrix}^{-1},
\end{align}
where we define $\mb B:=\mathbb{E}_{\mb x}(\mb x\mb x^T)+\sigma(t)^2\mb I$ for simplicity. By the Sherman–Morrison–Woodbury formula, we have:
\begin{align}
    \begin{bmatrix}
         \mb B &\mb \mu \\
         \mb \mu^T & 1
     \end{bmatrix}^{-1} = \begin{bmatrix}
         (\mb B-\mb \mu\mb \mu^T)^{-1} & -(\mb B-\mb \mu\mb \mu^T)^{-1}\mb \mu\\
         -(1-\mb \mu^T\mb B^{-1}\mb \mu)^{-1}\mb \mu^T\mb B^{-1} & (1-\mb \mu^T\mb B^{-1}\mb \mu)^{-1}
     \end{bmatrix}.
\end{align}
Therefore, we have:
\begin{align}
    \tilde{\mb W}\rightarrow \begin{bmatrix}
        \mathbb{E}_{\mb x}[\mb x\mb x^T](\mb B-\mb \mu\mb \mu^T)^{-1}-\frac{\mb \mu\mb \mu^T\mb B^{-1}}{1-\mb \mu^T\mb B^{-1}\mb \mu} & -\mathbb{E}_{\mb x}[\mb x \mb x^T](\mb B-\mb \mu\mb \mu^T)^{-1}\mb \mu+\frac{\mb \mu}{1-\mb \mu^T\mb B^{-1}\mb \mu}
    \end{bmatrix},
\end{align}
from which we have 
\begin{align}
    \mb W&\rightarrow \mathbb{E}_{\mb x}[\mb x\mb x^T](\mb B-\mb \mu\mb \mu^T)^{-1}-\frac{\mb \mu\mb \mu^T\mb B^{-1}}{1-\mb \mu^T\mb B^{-1}\mb \mu}\\
        \mb b&\rightarrow-\mathbb{E}_{\mb x}[\mb x \mb x^T](\mb B-\mb \mu\mb \mu^T)^{-1}\mb \mu+\frac{\mb \mu}{1-\mb \mu^T\mb B^{-1}\mb \mu}
\end{align}
Since $\mathbb{E}_{\mb x}[\mb x\mb x^T]=\mathbb{E}_{\mb x}[(\mb x-\mb \mu)((\mb x-\mb \mu)^T]+\mb \mu\mb \mu^T$, we have:
\begin{align}
    \mb W = \mb\Sigma(\mb\Sigma+\sigma(t)^2\mb I)^{-1}+\mb \mu\mb \mu^T(\mb B-\mb \mu\mb \mu^T)^{-1}-\frac{\mb \mu\mb \mu^T\mb B^{-1}}{1-\mb \mu^T\mb B^{-1}\mb \mu}.
\end{align}
Applying Sherman-Morrison Formula, we have:
\begin{align}
    (\mb B-\mb \mu\mb \mu^T)^{-1} = \mb B^{-1}+\frac{\mb B^{-1}\mb \mu\mb \mu^T\mb B^{-1}}{1-\mb \mu^T\mb B^{-1}\mb \mu},
\end{align}
therefore
\begin{align}
    \mb \mu\mb \mu^T(\mb B-\mb \mu\mb \mu^T)^{-1}-\frac{\mb \mu\mb \mu^T\mb B^{-1}}{1-\mb \mu^T\mb B^{-1}\mb \mu} &=\frac{\mb \mu\mb \mu^T\mb B^{-1}\mb \mu\mb \mu^T\mb B^{-1}}{1-\mb \mu^T\mb B^{-1}\mb \mu}-\frac{\mb \mu\mb \mu^T\mb B^{-1}\mb \mu^T\mb B^{-1}\mb \mu}{1-\mb \mu^T\mb B^{-1}\mb \mu}\\
    &=\frac{\mb \mu^T\mb B^{-1}\mb \mu}{1-\mb \mu^T\mb B^{-1}\mb \mu}(\mb \mu\mb \mu^T\mb B^{-1}-\mb \mu\mb \mu^T\mb B^{-1})\\
    &= \mb 0
\end{align}, which implies
\begin{align}
\mb W &\rightarrow \mb\Sigma(\mb\Sigma+\sigma(t)^2\mb I)^{-1}\\
    &=\mb U\Tilde{\mb\Lambda}_{\sigma(t)}\mb U^T.
\end{align}
Similarly, we have:
\begin{align}
    \mb b \rightarrow (\mb I-\mb U\Tilde{\mb\Lambda}_{\sigma(t)}\mb U^T)\mb \mu.
\end{align}
Therefore, gradient descent with a properly chosen learning rate $\eta$ recovers the Gaussian Denoisers when time goes to infinity.

\subsection{Two Extreme Cases}
Our empirical results indicate that the best linear approximation of $\mathcal{D}_{\mb \theta}$ is approximately equivalent to $\mc{D}_{\mathrm{G}}$.
According to the orthogonality principle~\cite{Hero2005STATISTICALMF}, this requires $\mathcal{D}_{\mb \theta}$ to satisfy:
    \begin{align}
    \label{orthogonal principle}
    \mathbb{E}_{\mb x\sim p_{\text{data}}(\mb x)}\mathbb{E}_{\mb \epsilon\sim \mathcal{N}(\mb 0;\sigma(t)^2\mb I)}\{(\mathcal{D}_{\mb\theta}(\mb x+\mb \epsilon;\sigma(t))-(\mb x-\mb \mu))(\mb x+\mb \epsilon-\mb\mu)^T\} \approx \mb 0.
\end{align}
Notice that~\eqref{orthogonal principle} does not hold for general denoisers. Two extreme cases for this to hold are:
\begin{itemize}
    \item Case 1: $\mathcal{D}_{\mb\theta}(\mb x+\mb \epsilon;\sigma(t))\approx \mb x$ for $~\forall\mb x \sim p_\text{data}, \mb \epsilon\sim\mathcal{N}(\mb 0,\sigma(t)^2\mb I)$.
    \item Case 2: $\mathcal{D}_{\mb\theta}(\mb x+\mb \epsilon;\sigma(t))\approx \mc{D}_{\mathrm{G}}(\mb x+\mb \epsilon;\sigma(t))$ for $~\forall\mb x \sim p_\text{data}, \mb \epsilon\sim\mathcal{N}(\mb 0,\sigma(t)^2\mb I)$.
\end{itemize}
Case 1 requires $\mathcal{D}_{\mb\theta}(\mb x+\mb \epsilon;\sigma(t))$ to be the oracle denoiser that perfectly recover the ground truth clean image, which never happens in practice except when $\sigma (t)$ becomes extremely small. Instead, our empirical results suggest diffusion models in the generalization regime bias towards Case 2, where deep networks learn $\mathcal{D}_{\mb\theta}$ that approximate (not equal) to $\mc{D}_{\mathrm{G}}$. This is evidenced in~\Cref{fig: uniqueness}(b), where diffusion models trained on larger datasets (35000 and 7000 images) produce denoising outputs similar to $\mc{D}_{\mathrm{G}}$. Notice that this similarity holds even when the denoisers take pure Gaussian noise as input. The exact mechanism driving diffusion models trained with gradient descent towards this particular solution remains an open question and we leave it as future work.

\section{More Discussion on Section 4}
While in~\Cref{sec: conditon for the inductive bias} we mainly focus on the discussion of the behavior of diffusion denoisers in the intermediate-noise regime, in this section we study the denoiser dynamics in both low and high-noise regime. We also provide additional experiment results on CIFAR-10 dataset. 

\label{additional results sec 4}
\subsection{Behaviors in Low-noise Regime} 
\label{Behaviors in low-noise regime}
\begin{figure}[b]
    \centering
    \includegraphics[width=1\linewidth]{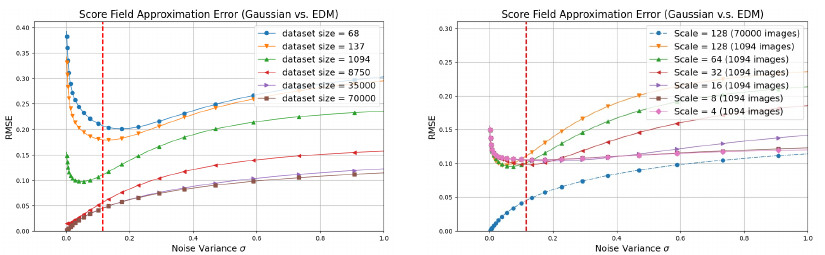}
    \caption{\textbf{Score differences for low-noise variances.} The left and right figures are the zoomed-in views of~\Cref{fig:various_dataset_size}(a) and~\Cref{fig:various_model_scale}(a) respectively. Notice that when the dataset size is smaller than the dimension of the image, the score differences are always non-zero at $\sigma=0$. }
    \label{fig:difference in low-noise reimge}
\end{figure}

We visualize the score differences between $\mc{D}_{\mathrm{G}}$ and $\mc D_{\mb \theta}$ in low-noise regime in~\Cref{fig:difference in low-noise reimge}. The left figure demonstrates that when the dataset size becomes smaller than a certain threshold, the score difference at $\sigma=0$ remains persistently non-zero. Moreover, the right figure shows that this difference depends solely on dataset size rather than model capacity. This phenomenon arises from two key factors:
\emph{(i)} $\mc D_{\mb \theta}$ converges to the identity mapping at low noise levels, independent of training dataset size and model capacity, and
\emph{(ii)} $\mc{D}_{\mathrm{G}}$ approximates the identity mapping at low noise levels only when the empirical covariance matrix is full-rank, as can be seen from~\eqref{Gaussian denoiser}.

Since the rank of the covariance matrix is upper-bounded by the training dataset size, $\mc{D}_{\mathrm{G}}$ differs from the identity mapping when the dataset size is smaller than the data dimension. This creates a persistent gap between $\mc{D}_{\mathrm{G}}$ and $\mc D_{\mb \theta}$, with smaller datasets leading to lower rank and consequently larger score differences. These observations align with our discussion in \Cref{sec: identity behavior}.

\subsection{Behaviors in High-noise Regime}
\label{sec: behavior in high-noise reigme}
\begin{figure}[t]
    \centering
    \vspace{-0.1cm}
    \includegraphics{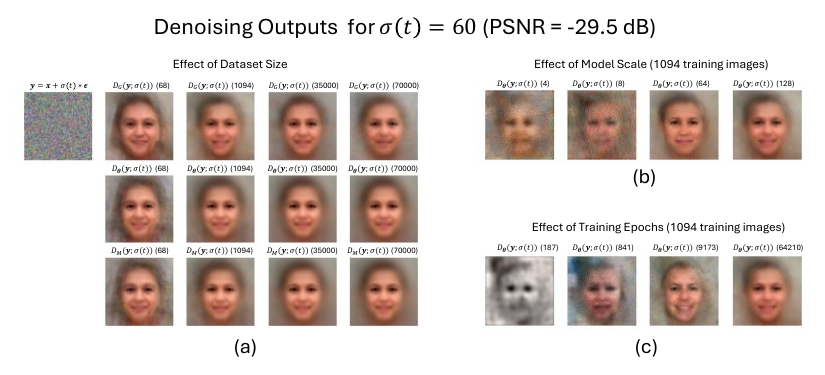}
    \caption{\textbf{$\mc D_{\mb \theta}$ converge to $\mc D_{\mathrm{G}}$ with no overfitting for high noise variances.} Figure(a) shows the denoising outputs of $\mathcal D_{\mathrm{M}}$, $\mc D_{\mathrm{G}}$ and well-trained (trained with sufficient model capacity till convergence) $\mc D_{\mb \theta}$. Notice that at high noise variance, the three different denoisers are approximately equivalent despite the training dataset size. Figure(b) shows the denoising outputs of $\mc D_{\mb \theta}$ with different model scales trained until convergence. Notice that $\mc D_{\mb \theta}$ converges to $\mc D_{\mathrm{G}}$ only when the model capacity is large enough. Figure(c) shows the denoising outputs of $\mc D_{\mb \theta}$ with sufficient large model capacity at different training epochs. Notice that $\mc D_{\mb \theta}$ converges to $\mc D_{\mathrm{G}}$ only when the training duration is long enough. } 
    \vspace{-0.1cm}
    \label{fig:High variance denoising output}
\end{figure}

As shown in~\Cref{fig:various_model_scale}(a), while a decreased model scale pushes $\mc D_{\mb \theta}$ in the intermediate noise region towards $\mc D_{\mathrm{G}}$, their differences enlarges in the high noise variance regime. This phenomenon arises because diffusion models employ a bell-shaped noise sampling distribution that prioritizes intermediate noise levels, resulting in insufficient training at high noise variances. A shown in~\Cref{fig:High variance denoising output}, for high $\sigma (t)$, $\mc D_{\mb \theta}$ converge to $\mc D_{\mathrm{G}}$ when trained with sufficient model capacity (\Cref{fig:High variance denoising output}(b)) and training time (\Cref{fig:High variance denoising output}(c)). This behavior is consistent irrespective of the training dataset sizes (\Cref{fig:High variance denoising output}(a)). Convergence in the high-noise variance regime is less crucial in practice, since diffusion steps in this regime contribute substantially less than those in the intermediate-noise variance regime—a phenomenon we analyze further in \Cref{RMSE vs NMSE metrics}.
\begin{figure}[t]
    \centering
    \includegraphics[width=1\linewidth]{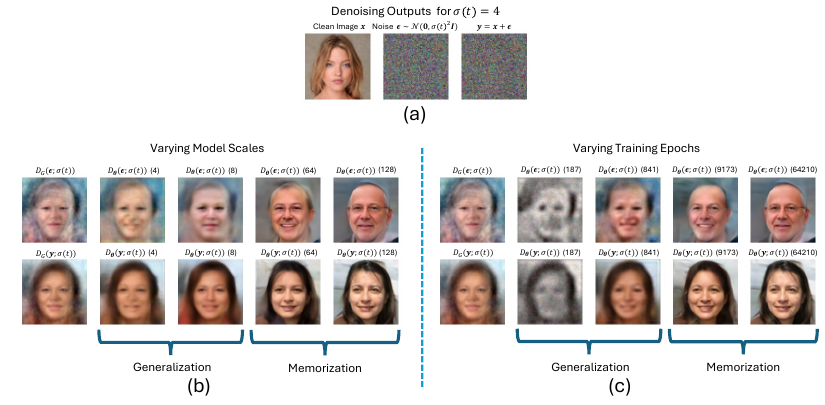}
    \caption{\textbf{Denoising outputs of $\mc D_{\mathrm{G}}$ and $\mc D_{\mb \theta}$ at $\sigma=4$.} Figure(a) shows the clean image $\mb x$ (from test set), random noise $\mb\epsilon$ and the resulting noisy image $\mb y$. Figure(b) compares denoising outputs of $\mc D_{\mb \theta}$ across different channel sizes [4, 8, 64, 128] with those of $\mc D_{\mathrm{G}}$. Figure(c) shows the evolution of $\mc D_{\mb \theta}$ outputs at training epochs [187, 841, 9173, 64210] alongside $\mc D_{\mathrm{G}}$ outputs. All models are trained on a fixed dataset of 1,094 images.}
    \label{fig:structural similarity}
\end{figure}

\subsection{Similarity between Diffusion Denoiers and Gaussian Denoisers}
In \Cref{sec: conditon for the inductive bias}, we demonstrate that the Gaussian inductive bias is most prominent in models with limited capacity and during early training stages, a finding qualitatively validated in \Cref{fig:structural similarity}. Specifically, \Cref{fig:structural similarity}(b) shows that larger models (channel sizes 128 and 64) tend to memorize, directly retrieving training data as denoising outputs. In contrast, smaller models (channel sizes 8 and 4) exhibit behavior similar to $\mc D_{\mathrm{G}}$, producing comparable denoising outputs. Similarly, \Cref{fig:structural similarity}(c) reveals that during early training epochs (0-841), $\mc D_{\mb \theta}$ outputs progressively align with those of $\mc D_{\mathrm{G}}$. However, extended training beyond this point leads to memorization.

\subsection{CIFAR-10 Results}
\begin{figure}[t]
    \centering
    \includegraphics[width=0.9\linewidth]{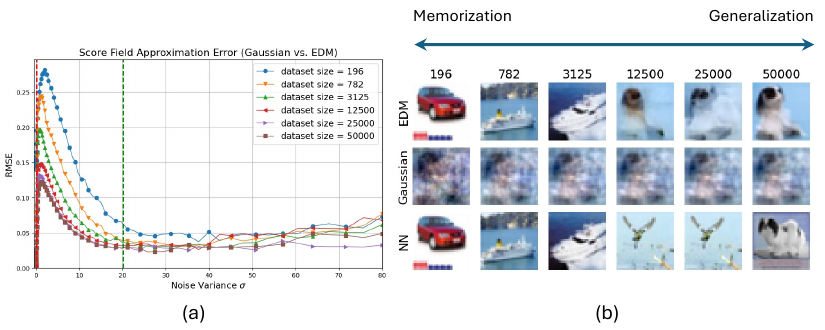}
    \caption{\textbf{Large dataset size prompts the Gaussian structure.} Models with the same scale (channel size 64) are trained on CIFAR-10 datasets with varying sizes. Figure(a) shows that larger dataset size leads to increased similarity between $\mc D_{\mathrm{G}}$ and $\mc D_{\mb \theta}$, resulting in structurally similar generated images as shown in Figure(b).}
    \label{fig:dataset size cifar}
\end{figure}

\begin{figure}[t]
    \centering
    \includegraphics[width=0.9\linewidth]{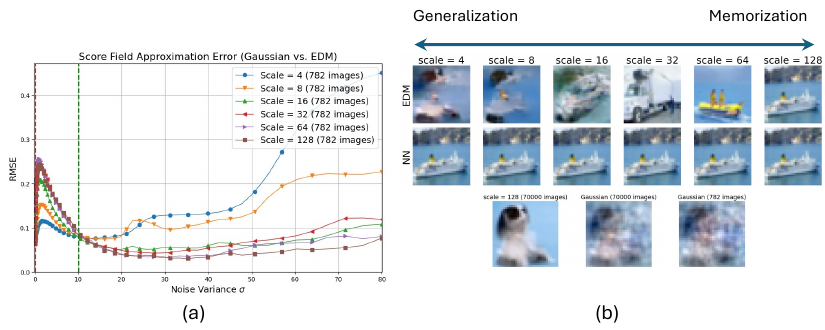}
    \caption{\textbf{Smaller model scale prompts the Gaussian structure.} Models with varying scales are trained on a fixed CIFAR-10 datasets with 782 images. Figure(a) shows that smaller model scale leads to increased similarity between $\mc D_{\mathrm{G}}$ and $\mc D_{\mb \theta}$ in the intermediate noise regime ($\sigma\in[0.1,10]$), resulting in structurally similar generated images as shown in figure(b). However, smaller scale leads to larger score differences in high-noise regime due to insufficient training from limited model capacity.}
    \label{fig:model scale cifar}
\end{figure}

\begin{figure}[t]
    \centering
    \includegraphics[width=0.9\linewidth]{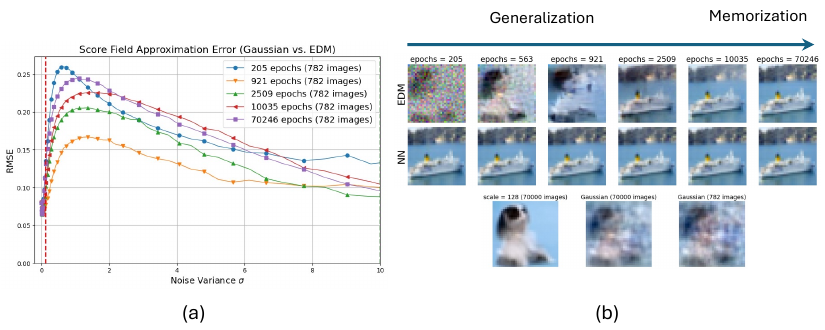}
    \caption{\textbf{Diffusion model learns the Gaussian structure in early training epochs.} Models with the same scale (channel size 128) are trained on a fixed CIFAR-10 datasets with 782 images. Figure(a) shows that the similarity between $\mc D_{\mathrm{G}}$ and $\mc D_{\mb \theta}$ progressively increases during early training epochs (0-921) in the intermediate noise regime ($\sigma\in[0.1,10]$), resulting in structurally similar generated images as shown in figure(b). However, continue training beyond this point results in diverged $\mc D_{\mathrm{G}}$ and $\mc D_{\mb \theta}$, resulting in memorization.}
    \label{fig:training epochs cifar}
\end{figure}

The effects of model capacity and training duration on the Gaussian inductive bias, as demonstrated in \Cref{fig:dataset size cifar,fig:model scale cifar,fig:training epochs cifar}, extend to the CIFAR-10 dataset. These results confirm our findings from \Cref{sec: conditon for the inductive bias}: the Gaussian inductive bias is most prominent when model scale and training duration are limited.

\section{Additional Experiment Results}
\label{Additional experiments}
While in the main text we mainly demonstrate our findings using EDM-VE diffusion models trained on FFHQ, in this section we show our results are robust and extend to various model architectures and datasets. Furthermore, we demonstrate that the Gaussian inductive bias is not unique to diffusion models, but it is a fundamental property of denoising autoencoders~\cite{vincent2008extracting}. Lastly, we verify that our conclusions remain consistent when using alternative metrics such as NMSE instead of the RMSE used in the main text.
\subsection{Gaussian Structure Emerges across Various 
Network Architectures}
\begin{figure}[t]
    \centering
    \includegraphics{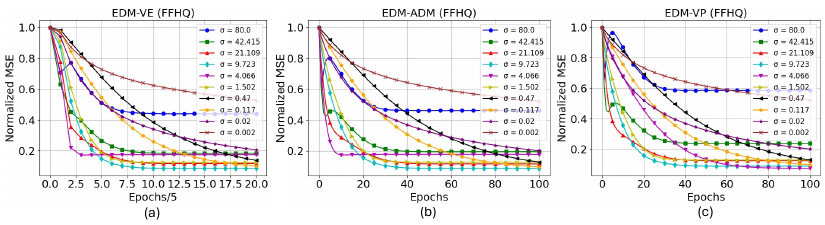}
    \caption{\textbf{Linear model shares similar function mapping with Gaussian model.} The figures demonstrate the evolution of normalized MSE between the linear weights $\mc D_{\mathrm{L}}$ and the Gaussian weights $\mc D_{\mathrm{G}}$ w.r.t. linear distillation training epochs. Figures(a), (b) and (c) correspond to diffusion models trained on FFHQ, with EDM-VE, EDM-ADM and EDM-VP network architectures specified in~\cite{karras2022elucidating} respectively.}
    \label{fig: learning curves linear distillation}
\end{figure}

\begin{figure}[t]
    \centering
    \includegraphics[width=0.6\textwidth]{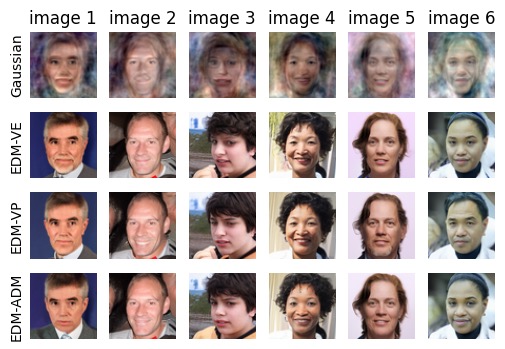}
    \caption{\textbf{Images sampled from various model.}The figure shows the sampled images from diffusion models with different network architectures and those from their corresponding Gaussian models.}
    \label{fig:reproducibility}
\end{figure}

We first demonstrate that diffusion models capture the Gaussian structure of the training dataset, irrespective of the deep network architectures used. As shown in \Cref{fig: learning curves linear distillation} (a), (b), and (c), although the actual diffusion models, $\mc D_{\mb\theta}$, are parameterized with different architectures, for all noise variances except $\sigma(t)\in\{0.002,80.0\}$, their corresponding linear models, $\mc D_{\mathrm{L}}$, consistently converge towards the common Gaussian models, \(\mc D_{\mathrm{G}}\), determined by the training dataset. Qualitatively, as depicted in \Cref{fig:reproducibility}, despite variations in network architectures, diffusion models generate nearly identical images, matching those generated from the Gaussian models. 

\subsection{Gaussian Inductive Bias as a General Property of DAEs}\label{sec: DAEs}
\label{sec: Gaussian inductive bias as a general property}
In previous sections, we explored the properties of diffusion models by interpreting them as collections of deep denoisers, which are equivalent to the denoising autoencoders (DAEs)~\cite{vincent2008extracting} trained on various noise variances by minimizing the denoising score matching objective~\eqref{Training Denoisers}. Although diffusion models and DAEs are equivalent in the sense that both of them are trying to learn the score function of the noise-mollified data distribution~\cite{vincent2011connection}, the training objective of diffusion models is more complex~\cite{karras2022elucidating}:
\begin{align}
    \min_{\mb\theta}\mathbb{E}_{\mb x,\mb \epsilon,\sigma}[\lambda(\sigma)c_\text{out}(\sigma)^2||\mathcal{F}_{\mb\theta}(\mb x+\mb \epsilon,\sigma)-\underbrace{\frac{1}{c_\text{out}(\sigma)}(\mb x-c_\text{skip}(\sigma)(\mb x+\mb \epsilon))}_{\text{linear combination of $\mb x$ and $\mb \epsilon$}}||_2^2],
\end{align}
where $\mb x\sim p_{\text{data}}, \mb \epsilon\sim\mathcal{N}(\mb 0,\sigma(t)^2\mb I)$ and $\sigma\sim p_{\text{train}}$.
Notice that the training objective of diffusion models has a few distinct characteristics:
\begin{itemize}
    \item Diffusion models use a single deep network $\mathcal{F}_{\mb \theta}$ to perform denoising score matching across all noise variances while DAEs are typically trained separately for each noise level.
    \item Diffusion models are not trained uniformly across all noise variances. Instead, during training the probability of sampling a given noise level $\sigma$ is controlled by a predefined distribution $p_{\text{train}}$ and the loss is weighted by $\lambda(\sigma)$. 
    \item Diffusion models often utilize special parameterizations~\eqref{denoiser parameterization}. Therefore, the deep network $\mathcal{F}_{\mb \theta}$ is trained to predict a linear combination of the clean image $\mb x$ and the noise $\mb \epsilon$, whereas DAEs typically predict the clean image directly.
\end{itemize}

Given these differences, we investigate whether the Gaussian inductive bias is unique to diffusion models or a general characteristic of DAEs. To this end, we train separate DAEs (deep denoisers) using the vanilla denoising score matching objective~\eqref{Training Denoisers} on each of the 10 discrete noise variances specified by the EDM schedule [80.0,
 42.415,
 21.108,
 9.723,
 4.06,
 1.501,
 0.469,
 0.116,
 0.020,
 0.002], and compare the score differences between them and the corresponding Gaussian denoisers $\mc D_{\mathrm{G}}$. We use no special parameterization so that $\mathcal{D}_{\mb\theta}=\mathcal{F}_{\mb\theta}$; that is, the deep network directly predicts the clean image. Furthermore, the DAEs for each noise variance are trained till convergence, ensuring all noise levels are trained sufficiently. We consider the following architectural choices:  
 
 \begin{itemize}
 \item \textit{DAE-NCSN}: In this setting, the network $\mathcal{F}_{\mb \theta}$ uses the NCSN architecture~\cite{songscore}, the same as that used in the EDM-VE diffusion model.
 \item \textit{DAE-Skip}: In this setting, $\mathcal{F}_{\mb \theta}$ is a U-Net~\cite{ronneberger2015u} consisting of convolutional layers, batch normalization~\cite{ioffe2015batch}, leaky ReLU activation~\cite{maas2013rectifier} and convolutional skip connections. We refer to this network as "Skip-Net".
Compared to NCSN, which adapts the state of the art architecture designs, Skip-Net is deliberately constructed to be as simple as possible to test how architectural complexity affects the Gaussian inductive bias.
 
 \item \textit{DAE-DiT}: In this setting, $\mathcal{F}_{\mb \theta}$ is a Diffusion Transformer (DiT) introduced in~\cite{peebles2023scalable}. Vision Transformers are known to lack inductive biases such as locality and translation equivariance that are inherent to convolutional models~\cite{dosovitskiy2020image}. Here we are interested in if this affects the Gaussian inductive bias.

 \item \textit{DAE-Linear}: In this setting we set $\mathcal{F}_{\mb \theta}$ to be a linear model with a bias term as in~\eqref{linear model with bias}. According to Theorem 1, these models should converge to Gaussian denoisers.
 \end{itemize}

\begin{figure}[t]
    \centering
    \includegraphics[width=1\linewidth]{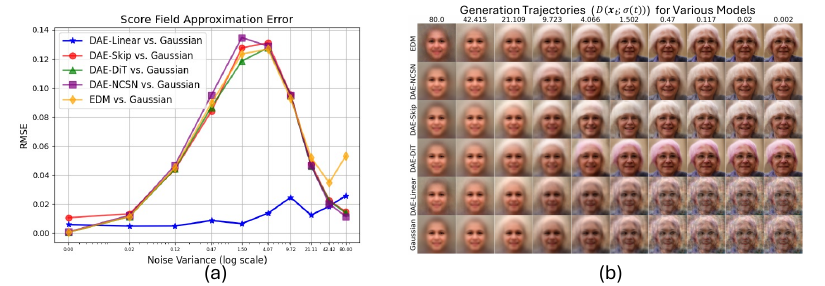}
    \caption{\textbf{Comparison between DAEs and diffusion models.} Figure(a) compares the score field approximation error between Gaussian models and both \emph{(i)} diffusion models (EDM vs. Gaussian) and \emph{(ii)} DAEs with varying architectures. Figure(b) illustrates the generation trajectories of different models initialized from the same noise input.}
    \label{fig:DAE v.s. Diffusion}
\end{figure}

The quantitative results are shown in~\Cref{fig:DAE v.s. Diffusion}(a). First, the DAE-linear models well approximate $\mc D_{\mathrm{G}}$ across all 10 discrete steps (RMSE smaller than 0.04), consistent with Theorem 1. Second, despite the differences between diffusion models (EDM) and DAEs, they achieve similar score approximation errors relative to $\mc D_{\mathrm{G}}$ for most noise variances, meaning that they can be similarly approximated by $\mc D_{\mathrm{G}}$. However, diffusion models exhibit significantly larger deviations from $\mc D_{\mathrm{G}}$ at higher noise variances ($\sigma \in \{42.415, 80.0\}$) since they utilize a bell-shaped noise sampling distribution $p_{\text{train}}$ that emphasizes training on intermediate noise levels, leading to under-training at high noise variances. Lastly, the DAEs with different architectures achieve comparable score approximation errors, and both DAEs and diffusion models generate images matching those from the Gaussian model, as shown in~\Cref{fig:DAE v.s. Diffusion}(b). These findings demonstrate that the Gaussian inductive bias is not unique to diffusion models or specific architectures but is a fundamental property of DAEs.

\begin{figure}[t!]
    \centering
    \includegraphics[width=1\textwidth]{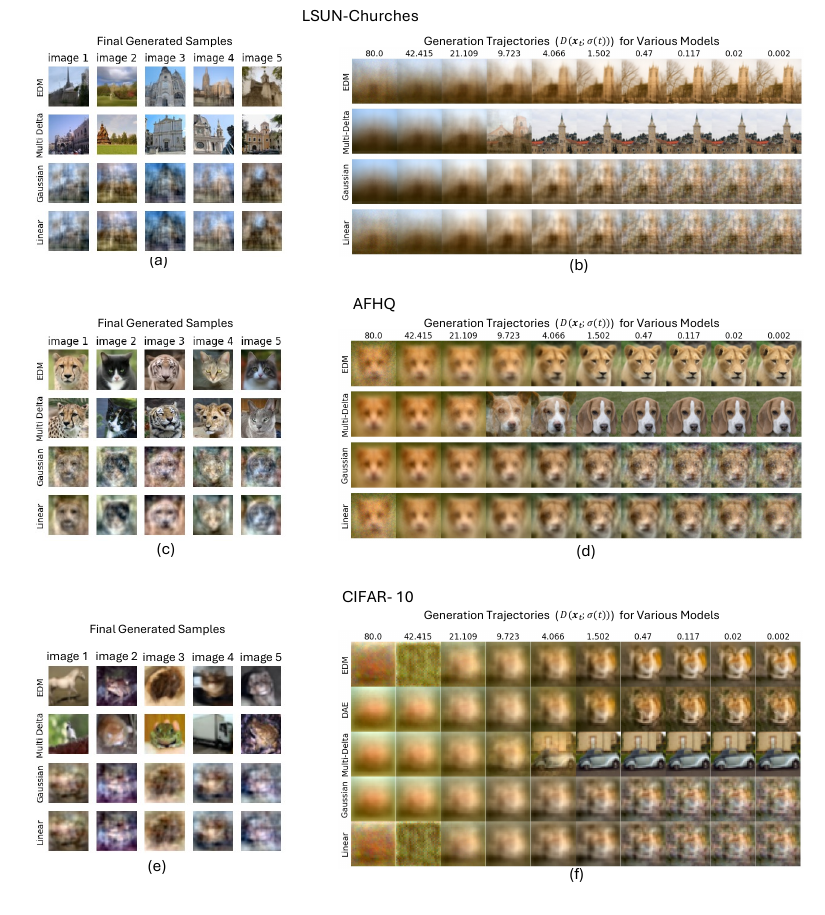}
    \caption{\textbf{Final generated images and sampling trajectories for various models.} Figures(a), (c) 
 and (e) demonstrate the images generated using different models starting from the same noises for LSUN-Churches, AFHQ and CIFAR-10 respectively. Figures(b), (d) and (f) demonstrate the corresponding sampling trajectories.}
    \label{fig: various datasets}
\end{figure}
\subsection{Gaussian Structure Emerges across Various datasets}
As illustrated in~\Cref{fig: various datasets}, for diffusion models trained on the CIFAR-10, AFHQ and LSUN-Churches datasets that are in the generalization regime, their generated samples match those produced by the corresponding Gaussian models. Additionally, their linear approximations, $\mc D_{\mathrm{L}}$, obtained through linear distillation, align closely with the Gaussian models, $\mc D_{\mathrm{G}}$, resulting in nearly identical generated images. These findings confirm that the Gaussian structure is prevalent across various datasets.

\subsection{Strong Generalization on CIFAR-10}
\begin{figure}[t]
    \centering
    \includegraphics[width=1\linewidth]{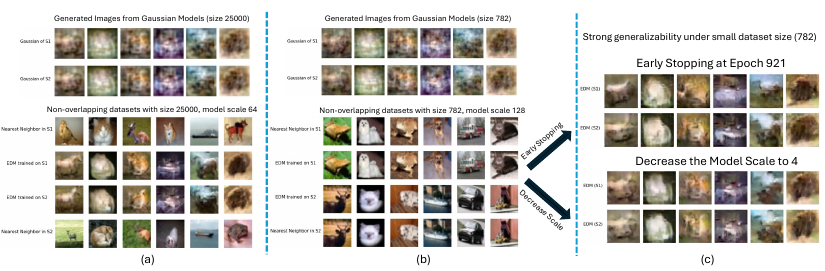}
     \caption{\textbf{Strong generalization on CIFAR-10 dataset.} Figure(a) Top: Generated images of Gausisan models; Bottom: Generated images of diffusion models, with model scale 64; S1 and S2 each has 25000 non-overlapping images. Figure(b) Top: Generated images of Gausisan model; Bottom: Generated images of diffusion models in the memorization regime, with model scale 128; S1 and S2 each has 782 non-overlapping images. Figure(c): Early stopping and reducing model capacity help transition diffusion models from memorization to generalization.}
    \label{fig:strong generalization cifar10}
\end{figure}

\Cref{fig:strong generalization cifar10} demonstrates the strong generalization effect on CIFAR-10. Similar to the observations in~\Cref{strong generalizability}, reducing model capacity or early stopping the training process prompts the Gaussian inductive bias, leading to generalization.

\subsection{Measuring Score Approximation Error with NMSE}
\label{RMSE vs NMSE metrics}
While in~\Cref{linear characteristic} we define the score field approximation error between denoisers $\mc D_1$ and $\mc D_2$ with RMSE (~\eqref{Score Error}), this error can also be quantified using NMSE:
\begin{align} \label{NMSE Score Error}
    \text{Score-Difference}(t):= \mathbb{E}_{\mb x\sim p_{\text{data}}(\mb x),\mb \epsilon\sim \mathcal{N}(\mb 0;\sigma(t)^2\mb I)}\frac{||\mc D_1(\mb x+\mb\epsilon)-\mc D_2(\mb x+\mb \epsilon)||_2}{||\mc D_1(\mb x+\mb\epsilon)||_2}.
\end{align}
\begin{figure}[t]
    \centering
    \includegraphics[width=0.9\linewidth]{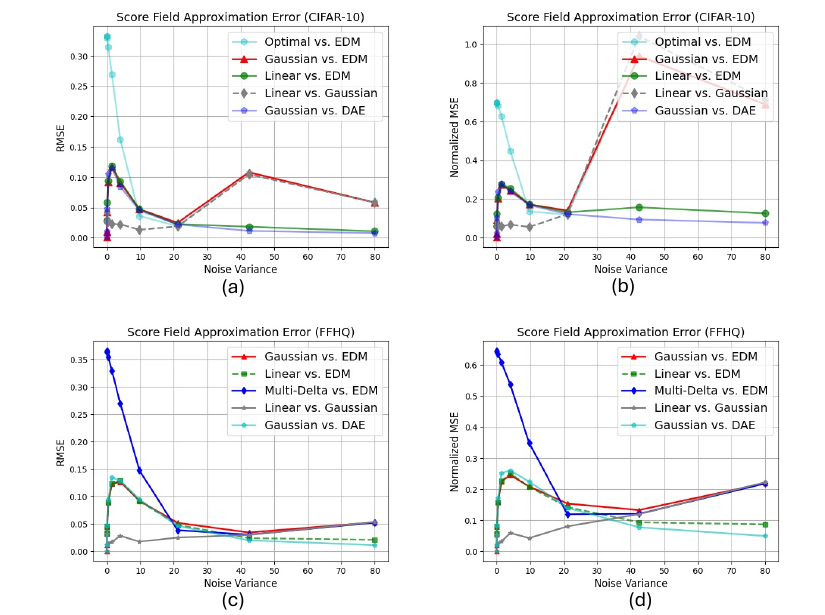}
    \caption{\textbf{Comparison between RMSE and NMSE score differences.} Figures(a) and (c) show the score field approximation errors measured with RMSE loss while figures(b) and (d) show these errors measured using NMSE loss. Compared to RMSE, the NMSE metric highlight the score differences in the high-noise regime, where diffusion models receive the least training.}
    \label{fig:RMSE vs NMSE}
\end{figure}
As shown in \Cref{fig:RMSE vs NMSE}, while the trend in intermediate-noise and low-noise regimes remains unchanged, NMSE amplifies differences in the high-noise variance regime compared to RMSE. This amplified score difference between $\mc D_{\mathrm{G}}$ and $\mc D_{\mb \theta}$ does not contradict our main finding that diffusion models in the generalization regime exhibit an inductive bias towards learning denoisers approximately equivalent to $\mc D_{\mathrm{G}}$ in the high-noise variance regime. As discussed in \Cref{inductive bias towards Gaussian,sec: behavior in high-noise reigme,sec: Gaussian inductive bias as a general property}, this large score difference stems from inadequate training in this regime.

\Cref{fig:RMSE vs NMSE} (Gaussian vs. DAE) demonstrates that when DAEs are sufficiently trained at specific noise variances, they still converge to $\mc D_{\mathrm{G}}$. Importantly, the insufficient training in the high-noise variance regime minimally affects final generation quality. \Cref{fig: various datasets}(f) shows that while the diffusion model (EDM) produces noisy trajectories at early timesteps ($\sigma\in\{80.0,42.415\}$), these artifacts quickly disappear in later stages, indicating that the Gaussian inductive bias is most influential in the intermediate-noise variance regime.

Notably, even when $\mc D_{\mb\theta}$ are inadequately trained in the high-noise variance regime, they remain approximable by linear functions, though these functions no longer match $\mc D_{\mathrm{G}}$.

\section{Discussion on Geometry-Adaptive Harmonic Bases}
\label{dicussion on GAHB}
\subsection{GAHB only Partially Explain the Strong Generalization}
Recent work~\cite{kadkhodaie2023generalization} observes that diffusion models trained on sufficiently large non-overlapping datasets (of the same class) generate nearly identical images. They explain this "strong generalization" phenomenon by analyzing bias-free deep diffusion denoisers with piecewise linear input-output mappings:
\begin{align}
    \mc D(\mb x_t;\sigma(t)) &= \nabla\mc D(\mb x_t;\sigma(t))\mb x \\
                           &= \sum_{k}\lambda_k(\mb x_t)\mb u_{k}(\mb x_t)\mb v_k^T(\mb x_t)\mb x_t,
\end{align}
where $\lambda_k(\mb x_t)$, $\mb u_k(\mb x_t)$, and $\mb v_k(\mb x_t)$ represent the input-dependent singular values, left and right singular vectors of the network Jacobian $\nabla\mc D(\mb x_t;\sigma(t))$. Under this framework, strong generalization occurs when two denoisers $\mc D_1$ and $\mc D_2$ have similar Jacobians: $\nabla\mc D_1(\mb x_t;\sigma(t))\approx \nabla\mc D_2(\mb x_t;\sigma(t))$. The authors conjecture this similarity arises from networks' inductive bias towards learning certain optimal $\nabla\mc D(\mb x_t;\sigma(t))$ that has sparse singular values and the singular vectors of which are the geometry-adaptive harmonic bases (GAHB)—near-optimal denoising bases that adapt to input $\mb x_t$.

While~\cite{kadkhodaie2023generalization} provides valuable insights, their bias-free assumption does not reflect real-world diffusion models, which inherently contain bias terms. For feed forward ReLU networks, the denoisers are piecewise affine:
\begin{align}
    \mc D(\mb x_t;\sigma(t)) &= \nabla\mc D(\mb x_t;\sigma(t))\mb x_t+\mb b_{\mb x_t},
\end{align}
where $\mb b_{\mb x_t}$ is the network bias that depends on both network parameterization and the noisy input $\mb x_t$~\cite{mohanrobust}. Here, similar Jacobians alone cannot explain strong generalization, as networks may differ significantly in $\mb b_{\mb x_t}$. For more complex network architectures where even piecewise affinity fails, we consider the local linear expansion of $\mc D(\mb x_t;\sigma(t))$:
\begin{align}
    \mc D(\mb x_t+\Delta\mb x;\sigma(t)) &= \nabla\mc D(\mb x_t;\sigma(t))\Delta\mb x_t+\mc D(\mb x_t;\sigma(t)),
\end{align}
which approximately holds for small perturbation $\Delta \mb x$. Thus, although $\nabla\mc D(\mb x_t;\sigma(t))$ characterizes $\mc D(\mb x_t;\sigma(t))$'s local behavior around $\mb x_t$, it does not provide sufficient information on the global properties.

Our work instead examines global behavior, demonstrating that $\mc D(\mb x_t;\sigma(t))$ is close to $\mc D_{\mathrm{G}}(\mb x_t;\sigma(t))$—the optimal linear denoiser under the Gaussian data assumption. This implies that strong generalization partially stems from networks learning similar Gaussian structures across non-overlapping datasets of the same class. Since our linear model captures global properties but not local characteristics, it complements the local analysis in~\cite{kadkhodaie2023generalization}.
\subsection{GAHB Emerge only in Intermediate-Noise Regime}
\begin{figure}[t]
    \centering
    \includegraphics[width=1\linewidth]{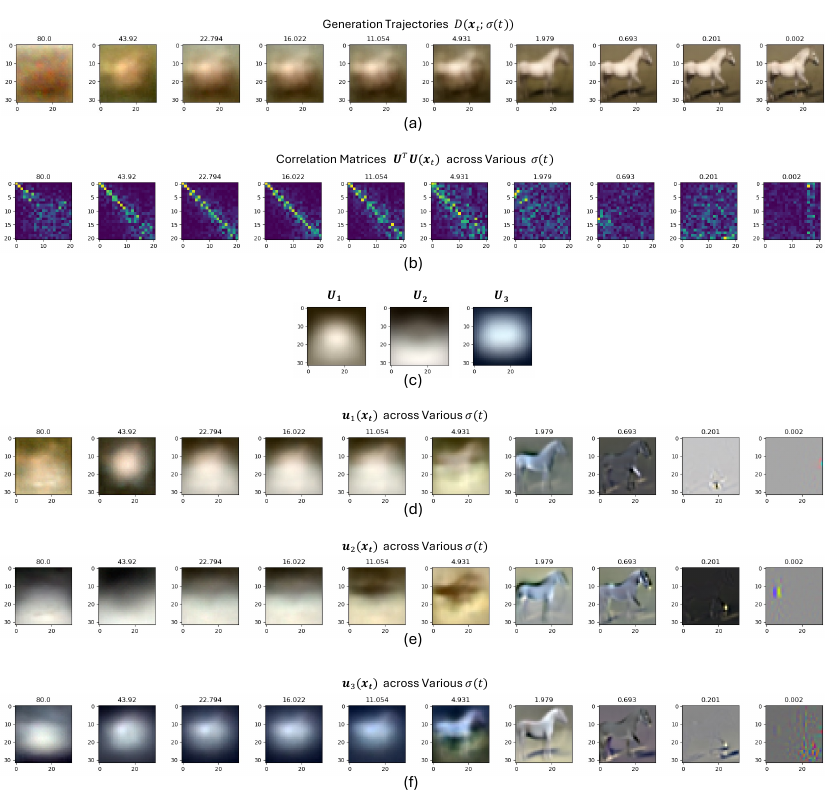}
    \caption{\textbf{Evolution of $\nabla\mc D(\mb x_t;\sigma(t))$ across varying noise levels.} Figure(a) shows the generation trajectory. Figure(b) shows the correlation matrix between Jacobian singular vectors $\mb U(\mb x_t)$ and training dataset principal components $\mb U$. Notice that the leading singular vectors of $\mb U(\mb x_t)$ and $\mb U$ well align in early timesteps but diverge in later timesteps. Figure(c) shows the first three principal components of the training dataset while figures(d-f) show the evolution of Jacobian's first three singular vectors across noise levels. These singular vectors initially match the principal components but progressively adapt to input image geometry, before losing distinct patterns at very low noise levels. While we present only left singular vectors, right singular vectors exhibit nearly identical behavior and yield equivalent results.}
    \label{fig:GAHB-EDM}
\end{figure}

\begin{figure}[t]
    \centering
    \includegraphics[width=1\linewidth]{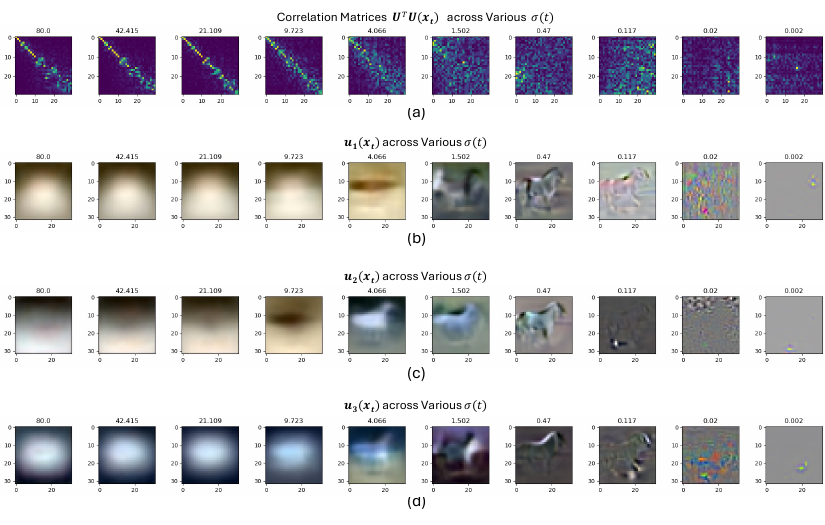}
    \caption{\textbf{Evolution of $\nabla\mc D(\mb x_t;\sigma(t))$ across varying noise levels for DAEs.} We repeat the experiments in~\Cref{fig:GAHB-EDM} on DAEs that are sufficiently trained on each discrete noise levels. Notice that with sufficient training, the Jacobian singular vectors $\mb U(\mb x_t)$ show a better alignment with principal components $\mb U$ in early timesteps.}
    \label{fig:GAHB-DAE}
\end{figure}

For completeness, we study the evolution of the Jacobian matrix $\nabla\mc D(\mb x_t;\sigma(t))$ across various noise levels $\sigma(t)$. The results are presented in~\Cref{fig:GAHB-EDM,fig:GAHB-DAE}, which reveal three distinct regimes:
\begin{itemize}[leftmargin=*]
    \item \emph{High-noise regime [10,80].} In this regime, the leading singular vectors\footnote{We only care about leading singular vectors since the Jacobians in this regime are highly low-rank. The less well aligned singular vectors have singular values near 0.} of the Jacobian matrix $\nabla\mc D(\mb x_t;\sigma(t))$ well align with those of the Gaussian weights (the leading principal components of the training dataset), consistent with our finding that diffusion denoisers approximate linear Gaussian denoisers in this regime. Notice that DAEs trained sufficiently on separate noise levels (\Cref{fig:GAHB-DAE}) show stronger alignment compared to vanilla diffusion models (\Cref{fig:GAHB-EDM}), which suffer from insufficient training at high noise levels.
    \item \emph{Intermediate-noise regime [0.1,10]:} In this regime, GAHB emerge as singular vectors of $\nabla\mc D(\mb x_t;\sigma(t))$ diverge from the principal components, becoming increasingly adaptive to the geometry of input image.
    
    \item \emph{Low-noise regime [0.002,0.1].} In this regime, the leading singular vectors of $\nabla\mc D(\mb x_t;\sigma(t))$ show no clear patterns, consistent with our observation that diffusion denoisers approach the identical mapping, which has unconstrained singular vectors.
\end{itemize}

\begin{figure}[t]
    \centering
    \includegraphics[width=1\linewidth]{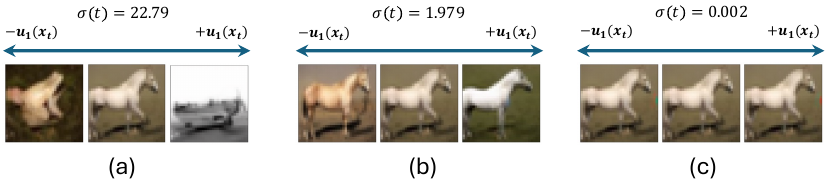}
    \caption{\textbf{Effects of perturbing $\mb x_t$ along Jacobian singular vectors.} Figure(a)-(c) demonstrate the effects of perturbing input $\mb x_t$ along the first singular vector of the Jacobian matrix ($\mb x_t \pm \lambda \mb u_1(\mb x_t)$) on the final generated images. Perturbing $\mb x_t$ in high-noise regime (Figure (a)) leads to canonical image changes while perturbation in intermediate-noise regime (Figure (b)) leads to change in details but the overall image structure is preserved. At very low noise variances, perturbation has no significant effect (Figure (c)). Similar effects are observed in concurrent work~\cite{chen2024exploring}.}
    \label{fig:Local-editing}
\end{figure}

Notice that the leading singular vectors of $\nabla\mc D(\mb x_t;\sigma(t))$ are the input directions that lead to the maximum variation in denoised outputs, thus revealing meaningful information on the local properties of $\mc D(\mb x_t;\sigma(t))$ at $\mb x_t$. 
As demonstrated in~\Cref{fig:Local-editing}, perturbing input $\mb x_t$ along these vectors at difference noise regimes leads to distinct effects on the final generated images: \emph{(i)} in the high-noise regime where the leading singular vectors align with the principal components of the training dataset, perturbing $\mb x_t$ along these directions leads to canonical changes such as image class, \emph{(ii)} in the intermediate-noise regime where the GAHB emerge, perturbing $\mb x_t$ along the leading singular vectors modify image details such as colors while preserving overall image structure and \emph{(iii)} in the low-noise regime where the leading singular vectors have no significant pattern, perturbing $\mb x_t$ along these directions yield no meaningful semantic changes.

These results collectively demonstrate that the singular vectors of the network Jacobian $\nabla\mc D(\mb x_t;\sigma(t))$ have distinct properties at different noise regimes, with GAHB emerging specifically in the intermediate regime. This characterization has significant implications for uncertainty quantification~\cite{manorposterior} and image editing~\cite{chen2024exploring}.

\section{Computing Resources}
\label{computing resources}
All the diffusion models in the experiments are trained on A100 GPUs provided by NCSA Delta GPU~\cite{boerner2023access}.

\clearpage
\newpage

\end{document}